\pdfoutput=1

\documentclass[11pt]{article}
\usepackage[final]{acl}

\usepackage{times}
\usepackage{latexsym}
\usepackage{hyperref}
\usepackage{url}
\usepackage{natbib}
\usepackage{booktabs}
\usepackage{multirow}
\usepackage{algorithm}
\usepackage{algpseudocode}
\usepackage{tikz-qtree}
\usepackage{array}
\usepackage{tabularx}
\usepackage{caption}
\usepackage{subcaption}
\usepackage{amssymb}

\let\emptyset\varnothing

\usepackage[T1]{fontenc}

\usepackage[utf8]{inputenc}

\usepackage{microtype}

\usepackage{inconsolata}

\usepackage{graphicx}
\newcommand{\AND}{\textbf{and} }

\newcommand\blfootnote[1]{%
  \begingroup
  \renewcommand\thefootnote{}\footnote{#1}%
  \addtocounter{footnote}{-1}%
  \endgroup
}

\title{BPE Gets Picky: Efficient Vocabulary Refinement \\ During Tokenizer Training}

\author{Pavel Chizhov*$^{1,2,3}$ ~~~ Catherine Arnett*$^{2,4}$ ~~~  Elizaveta Korotkova$^{3}$ ~~~ Ivan P. Yamshchikov$^{1,2}$ \vspace{2mm} \\
  $^{1}$Center for Artificial Intelligence, Technical University of Applied Sciences Würzburg-Schweinfurt \\
  $^{2}$PleIAs, Paris, France \\
  $^{3}$Institute of Computer Science, University of Tartu \\
  $^{4}$University of California, San Diego \\
  \texttt{\hphantom{pppp}pavel.chizhov@thws.de} \hphantom{pp} \texttt{catherine@pleias.fr\hphantom{pppppp}}  \\ \texttt{elizaveta.korotkova@ut.ee} \hphantom{pp} \texttt{ivan.yamshchikov@thws.de\hphantom{p}}} 

\begin{document}

\maketitle

\begin{abstract}
Language\blfootnote{*equal contribution} models can largely benefit from efficient tokenization. However, they still mostly utilize the classical BPE algorithm, a simple and reliable method. This has been shown to cause such issues as under-trained tokens and sub-optimal compression that may affect the downstream performance. We introduce Picky BPE, a modified BPE algorithm that carries out vocabulary refinement during tokenizer training. Our method improves vocabulary efficiency, eliminates under-trained tokens, and does not compromise text compression. Our experiments show that our method does not reduce the downstream performance, and in several cases improves it.
\end{abstract}

\section{Introduction} \label{intro}

Tokenization is a relatively understudied area, but it can greatly impact model performance and efficiency \citep{rust-etal-2021-good, hofmann2022embarrassingly, ali2023tokenizer, toraman2023impact, petrov2024language, singh2024tokenization, rajaraman2024toward, shao2024flexibly, wang2024tokenization}. Vocabularies should be efficient, as every additional token in the vocabulary increases embedding parameters, and thus model size. Each vocabulary item should contribute enough to model performance to justify the use of parameters. 

\begin{figure}[t!]
    \centering
    \includegraphics[width=\linewidth]{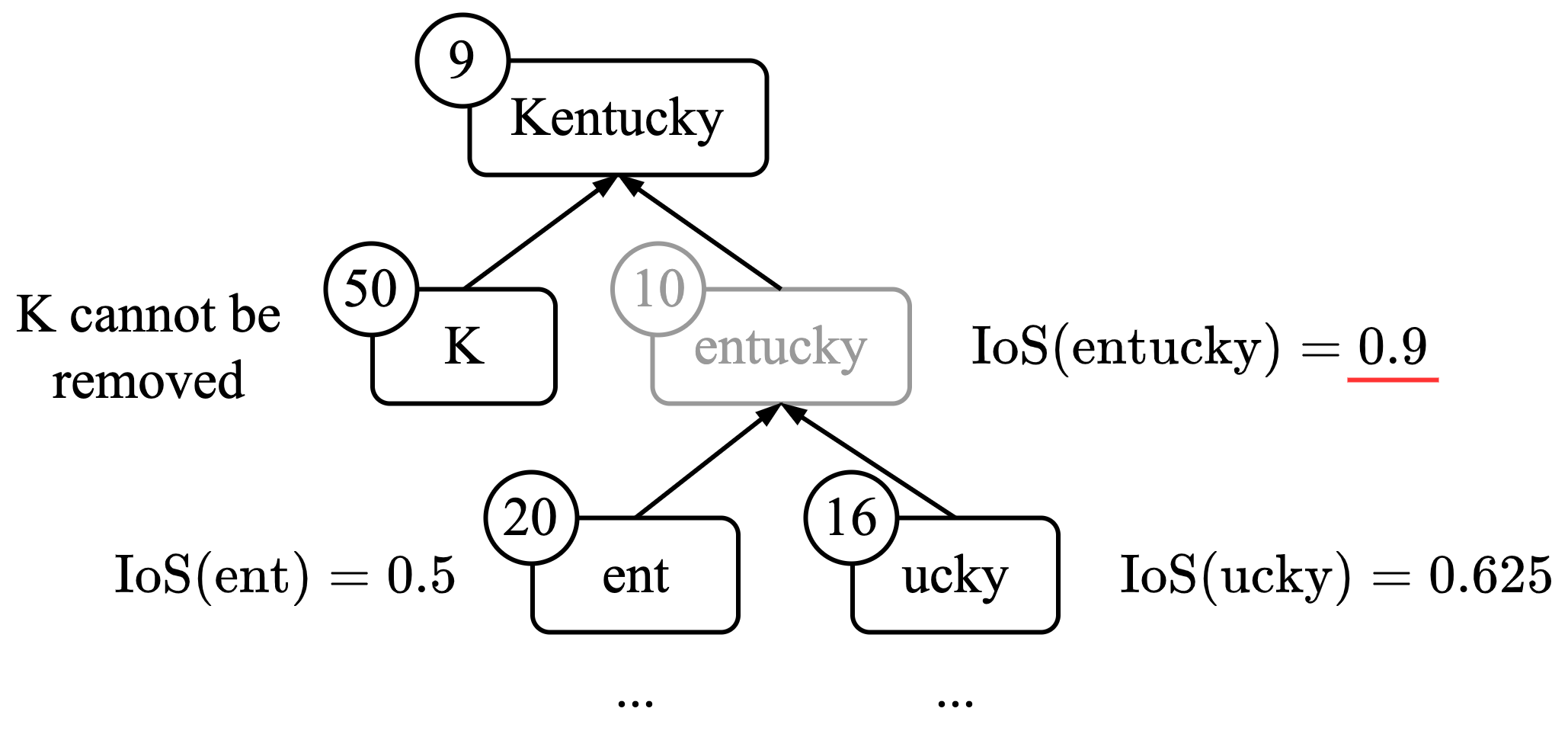}
    \caption{An example of a series of merges to produce a token \texttt{Kentucky}. The pre-merge token frequencies are demonstrated in corresponding circles. In the vanilla BPE algorithm, \texttt{entucky} should also be stored in the vocabulary, whereas it is redundant after the merge. In this example, the IoS metric effectively captures the intermediate token, as $\mathrm{IoS}(\texttt{entucky}) \geq \mathcal{T} = 0.9$.}
    \label{fig:kentucky}
\end{figure}

In this paper, we focus on Byte-Pair Encoding (BPE; \citet{gage1994new, sennrich-etal-2016-neural}) tokenizers. 
BPE tokenization works by breaking down a text into each of its characters or bytes and then building tokens in the vocabulary through a series of merges. The result of each merge must be stored as a token in the vocabulary. Tokens which are used only to execute merges are sometimes referred to as intermediate ``junk'' tokens \citep{bostrom-durrett-2020-byte}. An example is illustrated in Figure \ref{fig:kentucky}. Intermediate tokens clutter the vocabulary and are hardly ever used during tokenization. 

In addition to efficiency, we consider other model behaviors that may be driven by tokenization. \citet{land2024fishing} recently showed that very low-frequency tokens in the vocabulary may be under-trained by a model. This leads to worse downstream performance and unwanted outputs, such as hallucinations. Under-trained tokens can also potentially be exploited to avoid safety measures through the use of these out-of-distribution items. These tokens are also called ``glitch tokens'' \citep{rumbelow2023solidgoldmagikarp, geiping2024coercing, li2024glitch}.

Vocabulary trimming, which entails removing items from a tokenizer's vocabulary, has been proposed as a method to remove unnecessary tokens, \textit{e.g.}, language- or domain-specific tokens. Trimming has been shown to reduce embedding parameters without degrading downstream performance \citep{ushio-etal-2023-efficient, pang2024specialising}.  
Under-trained token indicators were shown to be correlated with token frequency in the training corpus, where less frequent tokens are more likely to be under-trained~\cite{land2024fishing}. Vocabulary trimming, thus, is well suited to address the issue of under-trained tokens.

Vocabulary trimming has mostly been implemented after tokenizer training \citep{yang-etal-2022-textpruner, cognetta2024analysis}. This means that it is difficult to determine the vocabulary size in advance because it is not known in advance how many tokens will be removed by the trimming procedure. Setting a fixed vocabulary size might be important, for example, in increasing training throughput  \citep{groeneveld2024olmo}.

In this paper, we introduce \textbf{Picky BPE}\footnote{\href{https://github.com/pchizhov/picky\_bpe}{https://github.com/pchizhov/picky\_bpe}} --- a modified BPE tokenizer that implements vocabulary refinement during tokenizer training. Unlike other trimming procedures, Picky BPE effectively removes intermediate tokens once they become useless and seamlessly collects the vocabulary of the desired size without data-specific heuristics. Our method leads to more efficient usage of the limited vocabulary, and thus the embedding parameters. We show that our method leads to equal or better performance on a downstream translation task (\S\ref{sec:translation}). Furthermore, we reduce the number of tokens that are likely to be under-trained (\S\ref{sec:under_trained}) and free space for higher-quality word-initial tokens. Due to the improved quality of the desired-size vocabulary, Picky BPE does not compromise text compression (\S\ref{sec:features}) unlike other trimming methods, which makes it suitable for practical use.

\section{Related Work}

Several common alternatives to BPE tokenization implicitly address the issue of intermediate low-frequency tokens. For instance, WordPiece tokenization~\citep{wu2016googles} is based on a series of merges akin to BPE, but along with the pair frequency, it also takes individual token frequencies into account. Thus, the tokenizer is less likely to add merges that would leave redundant tokens. However, this does not guarantee that the tokenizer adds merges in an optimal order, nor does it facilitate the retrospective removal of intermediate tokens that might eventually appear.

Another popular algorithm is Unigram tokenization \citep{kudo-2018-subword} used in SentencePiece~\citep{kudo-richardson-2018-sentencepiece}. The core of this algorithm is different from BPE-like solutions. Unigram works by creating a large vocabulary and iteratively pruning it until it reaches the desired size. The pruning is performed according to how much the token removal affects the likelihood of the subword sequence, and takes into account individual token frequencies. Intermediate tokens are also less likely to appear in such a scenario, which might suggest that Unigram tokenization implicitly performs a form of vocabulary trimming. We compare our method with Unigram tokenization in~\S\ref{sec:features}.

There are also several proposed modifications to BPE, which address the issues raised in \S\ref{intro}. 
BPE-Dropout was proposed to mitigate the issue of rare subwords by dropping merges randomly during tokenizer training \citep{provilkov-etal-2020-bpe}. This method regularizes the BPE training to expose a model to alternate tokenizations of the same strings, making it more robust to noisy input, such as misspellings. BPE-dropout also helps in reducing the under-training of low-frequency tokens. However, this method does not change the tokenizer vocabulary that is used during inference, and ultimately does not bear on the issue of vocabulary efficiency.

\citet{sennrich-etal-2017-university} use an absolute frequency cut-off to prevent very low-frequency tokens from being added to the vocabulary. 
Similarly, \citet{vilar-federico-2021-statistical} propose a stopping criterion in order to select the optimal vocabulary for BPE. The authors propose a maximum likelihood constraint, where
BPE stops adding merges during training if a merge decreases the overall likelihood of the token sequence.

\citet{cognetta2024analysis} implemented a vocabulary trimming method for BPE. They found that they were able to reduce the vocabulary size without significantly reducing downstream translation performance. Their method, however, did worsen compression. In some cases, when they showed the greatest task improvement, they found an increase of over 13\% in sequence length, \textit{i.e.}, text length in number of tokens. Better compression has been shown to correlate with better model performance  \citep{galle-2019-investigating, liang-etal-2023-xlm, goldman2024unpacking} and lead to faster inference time \citep{song-etal-2021-fast, petrov2024language, yamaguchi2024empirical}. 

Our Picky BPE differs from this method as we do not reduce the final vocabulary size. In addition, our trimming is performed during training, which preserves the overall distribution of token frequencies and does not require heuristic data-related post-processing, \textit{i.e.}, choosing an absolute frequency threshold that is different for every dataset~\cite{cognetta2024analysis}.

In a concurrent work, \citet{lian2024scaffold} also identify the issue of intermediate (``scaffold") tokens and introduce Scaffold-BPE. The authors propose to identify intermediate tokens when they are below the current range of frequencies during the tokenizer training. Compared to our method that uses a thresholding hyperparameter (see \S\ref{sec:alg}), there is no way to regulate the strength of Scaffold-BPE. In addition, the authors propose to run inference by first tokenizing the input text using both vocabulary and scaffold tokens and then splitting the scaffold tokens into the shortest valid token sequences. This approach does not strictly adhere to the training process and leads to inaccuracies in tokenization and worse compression (see the example and the comparison in Appendix~\ref{app:inference}).

Another contemporaneous work by \citet{bauwens-delobelle-2024-bpe} also proposes a method of pruning merges that lead to undesired segmentation and bloated vocabularies. This approach differs in at least two key ways from Picky BPE: 1) it allows merges of more than two tokens and 2) it uses a semi-supervised method to determine which merges to remove, based on manually annotated language-specific morphological segmentations.

\section{Picky BPE} \label{sec:alg}

\begin{algorithm}[t!]
\begin{algorithmic}[1]
\State \textbf{Input:} Vocabulary $\mathcal{V}$; Tokenized corpus $\mathcal{C}$;  Events order $\mathcal{E}$; IoS threshold $\mathcal{T}$
\State \textbf{Output:} Updated $\mathcal{V}, \mathcal{C}, \mathcal{E}$

\State $x_1, x_2 \leftarrow$ the most frequent pair in $\mathcal{C}$
\State $x_3 = x_1 
 + x_2$
\State $\mathcal{V} \leftarrow \mathcal{V} + \{x_3\}$
\State $\mathcal{E} \leftarrow \mathcal{E} + \{\mathrm{Merge}, (x_1, x_2)\}$ \Comment{new event}
\If {$\mathrm{IoS}(x_1\ |\ x_1, x_2) \ge \mathcal{T}$}
    \State $\mathcal{V} \leftarrow \mathcal{V} \setminus \{x_1\}$ \Comment{remove $x_1$}
    \State $\mathcal{E} \leftarrow \mathcal{E} + \{\mathrm{Remove}, x_1\}$ \Comment{new event}
\EndIf
\If {$x_2 \neq x_1$ \AND $\mathrm{IoS}(x_2\ |\ x_1, x_2) \ge \mathcal{T}$}
    \State $\mathcal{V} \leftarrow \mathcal{V} \setminus \{x_2\}$ \Comment{remove $x_2$}
    \State $\mathcal{E} \leftarrow \mathcal{E} + \{\mathrm{Remove}, x_2\}$ \Comment{new event}
\EndIf
\State Update $\mathcal{C}$ based on events from this iteration
\State \Return $\mathcal{V}, \mathcal{C}, \mathcal{E}$
\end{algorithmic}
\caption{Picky BPE Training Step}
\label{alg:merging}
\end{algorithm}

Our method is a modification of the original BPE algorithm~\citep{gage1994new, sennrich-etal-2016-neural}. The intuition behind our modification is that we can identify intermediate tokens based on their individual frequency and frequency within a larger token. 
Intermediate tokens should have low frequency outside of the context of the token that contains them. For example, in Figure~\ref{fig:kentucky}, an intermediate token \texttt{entucky} is almost always a part of \texttt{Kentucky}, which is easy to capture by comparing the frequencies of \texttt{Kentucky} and \texttt{entucky}. To formalize this approach, we introduce a measure called \textit{Intersection over Self (IoS)}, which is computed as follows: 

\begin{equation}
    \mathrm{IoS}(x_1\ |\ x_1, x_2) = \frac{f_p(x_1, x_2)}{f_t(x_1)};
\end{equation}

\begin{equation}
    \mathrm{IoS}(x_2\ |\ x_1, x_2) = \frac{f_p(x_1, x_2)}{f_t(x_2)}.
\end{equation}

Here $x_1$ and $x_2$ are the tokens being merged, $f_t$ is token frequency, and $f_p$ is pair frequency. 
% IoS metric is also illustrated in Figure~\ref{fig:ios}. 
$\mathrm{IoS}(x_1\ |\ x_1, x_2)$ shows how often token $x_1$ occurs as part of a pair $\{x_1, x_2\}$ compared to all occurrences of $x_1$. If this value is too high, \textit{i. e.}, close to 1, $x_1$ is highly likely an intermediate token, an integral part of a longer, more meaningful token $x_1 + x_2$. Adding $x_1 + x_2$ to the vocabulary makes $x_1$ redundant and we can consider removing it. 

\begin{algorithm}[t!]
\begin{algorithmic}[1]
\State \textbf{Input:} Word $w$; Vocabulary $\mathcal{V}$; Events order $\mathcal{E}$
\State \textbf{Output:} Tokenized word $\mathcal{W}$

\State $\mathcal{W} \leftarrow$ split $w$ into symbols $\in \mathcal{V}$ 
\State $\mathcal{M} \leftarrow$ possible merges in $\mathcal{W}$
\State $\mathcal{R} \leftarrow$ possible removals in $\mathcal{W}$
\While{$\mathcal{M} \neq \emptyset$ \textbf{or} $\mathcal{R} \neq \emptyset$}
    \State $\varepsilon \leftarrow$ earliest event in $\mathcal{E}$, $\varepsilon \in \mathcal{M} \cup \mathcal{R}$
    \State perform $\varepsilon$
    \State update $\mathcal{M}, \mathcal{R}$
    \State exclude events from $\mathcal{E}$ earlier than $\varepsilon$
\EndWhile
\State \Return $\mathcal{W}$
\end{algorithmic}
\caption{Picky BPE Tokenization}
\label{alg:tokenization}
\end{algorithm}

\begin{figure*}[!htbp]
    \centering
    \includegraphics[width=\textwidth]{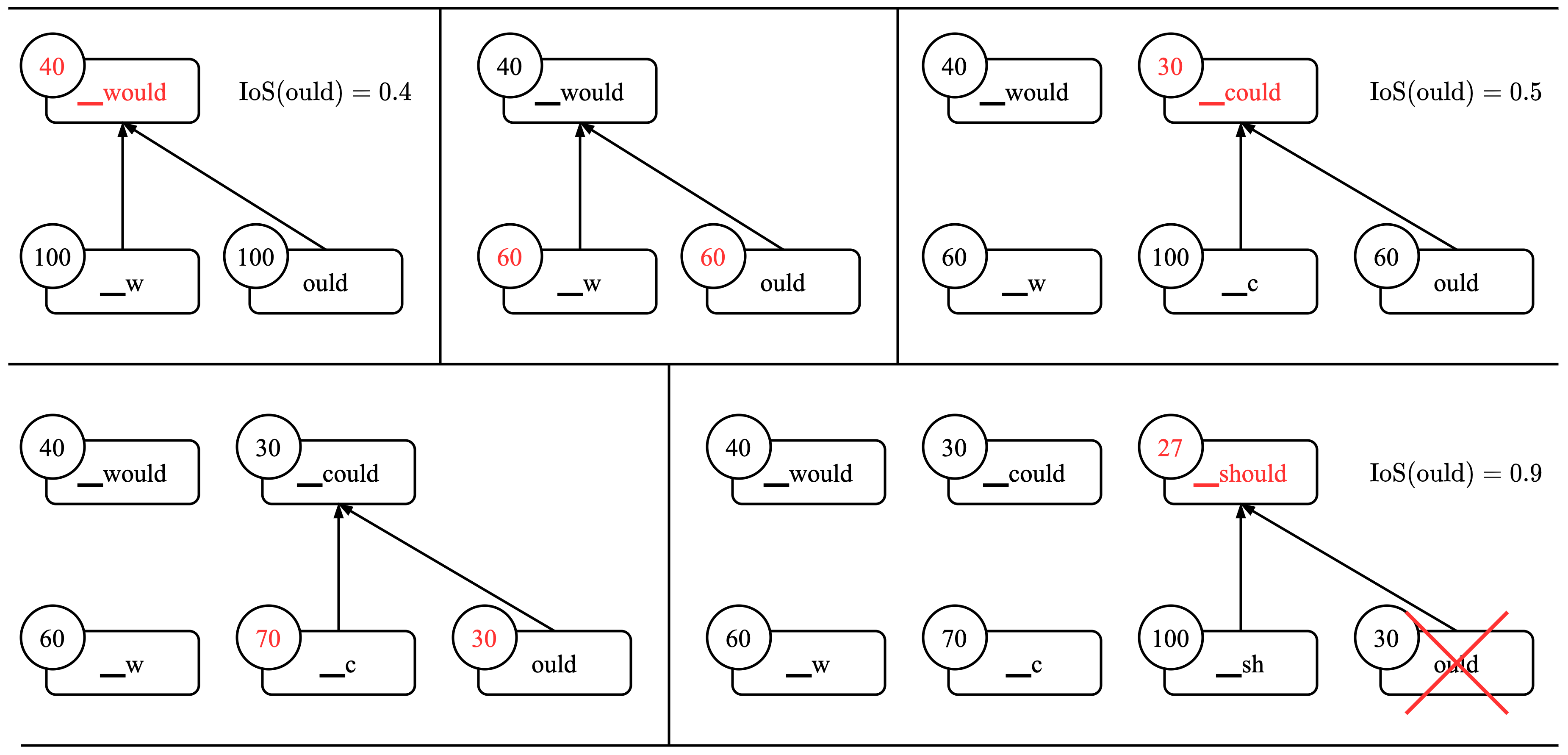}
    \caption{Picky BPE tokenization example. Token frequencies are demonstrated in the corresponding circles and are updated on merges. Token \texttt{``ould''} is removed only after merging into three common tokens containing it. The corresponding IoS values are visualized on every merge. Once IoS becomes greater or equal to the threshold $\mathcal{T}$, 0.9 in this example, the token \texttt{``ould''} is removed.}
    \label{fig:should_would_could}
\end{figure*}

\subsection{Algorithm description} \label{sec:algorithm_description}

The training of Picky BPE follows the main steps of the vanilla BPE training. The text is first split into a sequence of characters/bytes, initializing the vocabulary with unique symbols. Optionally, the coverage parameter (we use 0.9999 in our experiments) is used to replace the rarest symbols with \texttt{<unk>}. After that, the algorithm iteratively chooses the most frequent pair of tokens to merge and adds it to the vocabulary. The difference comes after each merge when we check whether we can remove any of the merged tokens judging by the IoS value. The pseudocode for a training step is demonstrated in Algorithm~\ref{alg:merging}. 
We integrate the IoS metric into the merging stage. When a pair of tokens is merged, we check whether we can safely remove either of the two tokens from the vocabulary. For this, we introduce a hyperparameter $\mathcal{T}$, the IoS threshold. If $\mathrm{IoS}(x_1\ |\ x_1,x_2) \geq \mathcal{T}$, we remove $x_1$. Thus, $\mathcal{T}$ leverages the strength of the removal policy: $\mathcal{T}$ is a positive value $\leq1$, and the closer it is to 1, the less strict becomes the removing criterion. For instance, $\mathcal{T}=0.9$ means that only the tokens present not more than 10\% of the time outside of the current merge will be removed. In an extreme case, $\mathcal{T}=1$ means that no removals are possible, thus the algorithm becomes the vanilla BPE.  
Another unique feature of our algorithm is that the merges and removals are stored in the events order array $\mathcal{E}$ in the order of happening. The events order is crucial for the tokenization step.

The tokenization (inference) step is described in Algorithm~\ref{alg:tokenization}. We first split the input word into a series of in-vocabulary symbols. Then we collect the sets of possible merges and removals in the current tokenization and iteratively greedily choose the earliest possible event using event order $\mathcal{E}$. The action associated with the chosen event is performed and the sets of possible merges and removal are updated. This process strictly follows the tokenizer training and avoids compression issues happening in the approximation methods (see Appendix~\ref{app:inference}).

\subsection{Algorithm analysis and justification}
The training of Picky BPE is longer than that of the original vanilla BPE. However, the difference is not drastic. When a token is removed, recalculating the distances requires a constant number of operations, which makes the training time depend linearly on the number of events (merges and removals). With threshold $\mathcal{T}$ values of 0.6 and higher, the proportion of removed tokens generally does not surpass 10\% (for details refer to Appendix~\ref{app:removed}), which makes the number of removals inferior to the number of merges. On the tokenization stage, the time depends on the number of evens, just as the tokenization time of the vanilla BPE depends on the number of merges. As we show in Appendix~\ref{app:removed}, merges comprise the largest partition of overall events, therefore removal events do not significantly slow down the inference. Depending on the programming language and the implementation, the astronomical time of both stages of the algorithm can differ significantly.

Here we also enumerate several algorithmic advantages of the proposed method.

\paragraph{Universal threshold.} The threshold $\mathcal{T}$ is relative and does not depend on the size of the training corpus or the desired vocabulary. This is one of the advantages of our method compared to the main counterparts, such as~\citet{cognetta2024analysis}. Furthermore, the removals happen during training resulting in the desired vocabulary size that does not require any post-processing.

\paragraph{Variety of intermediate tokens.} An intermediate token may be part of more than one token, as shown in Figure~\ref{fig:should_would_could}. Our algorithm handles these cases, removing the token only after there are few to no words it can be merged into.

\paragraph{Second chances.} Any removed token may be merged again if its frequency is higher at a later point in the order of merges. 
This is usually the case for tokens removed in the very beginning when the frequencies of new tokens are very high. For example, \texttt{(``t'', ``he'')} is likely to be merged early in tokenizer training because \texttt{``the''} is a frequent word. Because the relative frequency of \texttt{``he''} is lower, \texttt{``he''} may be split into \texttt{(``h'', ``e'')}. But because \texttt{``he''} is still a high-frequency word, it is likely to be merged again. If a previously removed token is restored, it is re-activated to keep its original place in the list of merges. This is essential to the merge order during tokenization.

\section{Machine Translation Experiments} \label{sec:translation}

To evaluate the downstream performance of our algorithm, we conduct several machine translation (MT) experiments.
We experiment with three translation directions: English--German (EN--DE), German--Estonian (DE--ET), and Ukrainian--Estonian (UK--ET). With this choice of language pairs, we aim to cover diverse MT tasks of varying difficulty. German and English are related languages and share the same script. This language pair represents an easier translation task. German and Estonian use the same script, but are much less closely related, belonging to different language families. Translation for this pair should be more difficult. Finally, Ukrainian and Estonian represent the most difficult translation pair in our experiments. These languages are not only distant but also use different scripts.

To train the EN--DE models, we use the training corpus from the WMT16 news translation task~\cite{Bojar2016FindingsOT}, with \texttt{newstest2016} corpus for evaluation. For DE--ET and UK--ET, we use the mixtures of parallel corpora assembled by~\citet{estonian_centric_mt}. For the evaluations of outputs in Estonian, we use the development set of the \textsc{Flores} benchmark~\cite{10.1162/tacl_a_00474}.

\begin{table}[t!]
    \centering
    \begin{tabular}{lccc}
    \toprule
        Experiment & $\mathcal{T}$\hphantom{$^{\ast}$} & BLEU $(\uparrow)$ & COMET $(\uparrow)$ \\
        \midrule
        \multirow{5}{*}{EN--DE} & 1.0$^{\ast}$ & 30.1 $\pm$ 0.7 & 0.431 \\
        \cmidrule{2-4}
        & 0.9\hphantom{$^{\ast}$} & 30.3 $\pm$ 0.7 & 0.431 \\
        & 0.8\hphantom{$^{\ast}$} & 30.0 $\pm$ 0.7 & 0.431 \\
        & 0.7\hphantom{$^{\ast}$} & 30.6 $\pm$ 0.7 & \textbf{0.434} \\
        & 0.6\hphantom{$^{\ast}$} & 30.3 $\pm$ 0.7 & 0.431 \\
        \midrule
        \multirow{5}{*}{DE--ET} & 1.0$^{\ast}$ & 19.4 $\pm$ 1.0 & 0.516 \\
        \cmidrule{2-4}
        & 0.9\hphantom{$^{\ast}$} & 19.9 $\pm$ 1.0 & \textbf{0.520} \\
        & 0.8\hphantom{$^{\ast}$} & 19.8 $\pm$ 1.0 & \textbf{0.520} \\
        & 0.7\hphantom{$^{\ast}$} & 19.9 $\pm$ 1.0 & \textbf{0.520} \\
        & 0.6\hphantom{$^{\ast}$} & 19.9 $\pm$ 1.1 & \textbf{0.520} \\
        \midrule
        \multirow{5}{*}{UK--ET} & 1.0$^{\ast}$ & 16.9 $\pm$ 1.0 & 0.506 \\
        \cmidrule{2-4}
        & 0.9\hphantom{$^{\ast}$} & 15.8 $\pm$ 1.5 & 0.508 \\
        & 0.8\hphantom{$^{\ast}$} & 16.7 $\pm$ 1.3 & \textbf{0.511} \\
        & 0.7\hphantom{$^{\ast}$} & 17.2 $\pm$ 1.0 & 0.509 \\
        & 0.6\hphantom{$^{\ast}$} & 16.9 $\pm$ 0.9 & \textbf{0.511} \\
    \bottomrule
    \end{tabular}
    \caption{Machine translation results with vocabulary size 8192 on \texttt{newstest2016} set~\cite{Bojar2016FindingsOT} for EN--DE, and on \textsc{Flores}-dev~\cite{10.1162/tacl_a_00474} for DE--ET and UK--ET. For every threshold~$\mathcal{T}$, we report BLEU~\citep{papineni-etal-2002-bleu} and COMET~\citep{rei-etal-2020-comet} scores on the translation task. The best scores are highlighted in \textbf{bold}. Other scores that are not statistically significantly different from the best are also highlighted in \textbf{bold}. If neither of the scores is significantly better, nothing is highlighted. $^{\ast}\mathcal{T} = 1.0$ represents the baseline vanilla BPE without intermediate token removal.}
    \label{tab:mt}
\end{table}

We test our method with several thresholds: 0.6, 0.7, 0.8, 0.9. We did not consider lower thresholds as they would remove too many useful tokens.
For the baseline, we chose vanilla BPE, which we obtained by training our Picky BPE with $\mathcal{T} = 1$ to ensure effects are not driven by implementation differences. We use the \texttt{transformer-iwslt} model from \texttt{fairseq}~\cite{ott-etal-2019-fairseq} for all translation tasks. The architecture and training details can be found in Appendix~\ref{app:arch}.

For generation, we use beam search with beam size 5 in all our experiments. We use BLEU~\cite{papineni-etal-2002-bleu} from \texttt{sacreBLEU}~\cite{post-2018-call} and COMET~\cite{rei-etal-2020-comet} scores for automatic evaluation. We compute paired t-Test with bootstrapping\footnote{We evaluate 1000 bootstrap resamples and use t-Test with confidence level 0.95.} to compare the obtained translation systems with statistical significance~\cite{koehn-2004-statistical}. 

\paragraph{Smaller vocabularies.} 
First, we conduct experiments on all three language pairs with a small vocabulary size of 8192. 
We chose such a restrictive setting to make sure all the tokens are sufficiently trained, as the relatively small training datasets we used ($\sim$1--4M sentence pairs) do not necessitate large vocabularies \citep{sennrich-zhang-2019-revisiting} and the effect of using our method might be less pronounced. The results are presented in Table~\ref{tab:mt}. 
Overall, the models trained with Picky BPE vocabulary performed comparably to the vanilla BPE, with at least one Picky BPE threshold significantly outperforming it for all three translation directions according to the COMET metric. COMET scores for the DE--ET experiment show that all Picky BPE models were better than the vanilla baseline.

\begin{table}[!t]
    \centering
    \begin{tabular}{cccc}
    \toprule
        Vocabulary & $\mathcal{T}$\hphantom{$^{\ast}$} & BLEU $(\uparrow)$ & COMET $(\uparrow)$ \\
        \midrule
        \multirow{5}{*}{\shortstack{8192\\+\\8192}} & 1.0$^{\ast}$ & 30.7 $\pm$ 0.7 & 0.431 \\
        \cmidrule{2-4}
        & 0.9\hphantom{$^{\ast}$} & 30.4 $\pm$ 0.7 & 0.431 \\
        & 0.8\hphantom{$^{\ast}$} & 30.3 $\pm$ 0.7 & 0.430 \\
        & 0.7\hphantom{$^{\ast}$} & 30.3 $\pm$ 0.7 & 0.430 \\
        & 0.6\hphantom{$^{\ast}$} & 30.8 $\pm$ 0.7 & \textbf{0.432} \\
        \midrule
        \multirow{5}{*}{\shortstack{16384\\+\\16384}} & 1.0$^{\ast}$ & 31.1 $\pm$ 0.7 & 0.433 \\
        \cmidrule{2-4}
        & 0.9\hphantom{$^{\ast}$} & 31.1 $\pm$ 0.7 & 0.433 \\
        & 0.8\hphantom{$^{\ast}$} & 31.0 $\pm$ 0.7 & \textbf{0.435} \\
        & 0.7\hphantom{$^{\ast}$} & 31.4 $\pm$ 0.7 & \textbf{0.435} \\
        & 0.6\hphantom{$^{\ast}$} & 31.1 $\pm$ 0.7 & \textbf{0.435} \\
        \midrule
        \multirow{5}{*}{\shortstack{32768\\+\\32768}} & 1.0$^{\ast}$ & 29.8 $\pm$ 0.7 & 0.418 \\
        \cmidrule{2-4}
        & 0.9\hphantom{$^{\ast}$} & 29.6 $\pm$ 0.8 & \textbf{0.428} \\
        & 0.8\hphantom{$^{\ast}$} & 30.5 $\pm$ 0.7 & \textbf{0.430} \\
        & 0.7\hphantom{$^{\ast}$} & 30.4 $\pm$ 0.7 & \textbf{0.430} \\
        & 0.6\hphantom{$^{\ast}$} & 28.3 $\pm$ 0.8 & 0.416 \\
        \midrule
        \multirow{5}{*}{16384} & 1.0$^{\ast}$ & 31.1 $\pm$ 0.7 & 0.436 \\
        \cmidrule{2-4}
        & 0.9\hphantom{$^{\ast}$} & 31.2 $\pm$ 0.7 & 0.436 \\
        & 0.8\hphantom{$^{\ast}$} & 30.9 $\pm$ 0.6 & 0.434 \\
        & 0.7\hphantom{$^{\ast}$} & 31.1 $\pm$ 0.7 & 0.436 \\
        & 0.6\hphantom{$^{\ast}$} & 31.3 $\pm$ 0.7 & \textbf{0.438} \\
        \midrule
        \multirow{5}{*}{32768} & 1.0$^{\ast}$ & 30.9 $\pm$ 0.7 & \textbf{0.435} \\
        \cmidrule{2-4}
        & 0.9\hphantom{$^{\ast}$} & 31.1 $\pm$ 0.7 & 0.434 \\
        & 0.8\hphantom{$^{\ast}$} & 31.1 $\pm$ 0.7 & \textbf{0.437} \\
        & 0.7\hphantom{$^{\ast}$} & 30.9 $\pm$ 0.7 & \textbf{0.436} \\
        & 0.6\hphantom{$^{\ast}$} & 30.9 $\pm$ 0.7 & 0.431 \\
        \midrule
        \multirow{5}{*}{65536} & 1.0$^{\ast}$ & 28.5 $\pm$ 0.7 & 0.421 \\
        \cmidrule{2-4}
        & 0.9\hphantom{$^{\ast}$} & 28.4 $\pm$ 0.7 & \textbf{0.427} \\
        & 0.8\hphantom{$^{\ast}$} & 28.6 $\pm$ 0.7 & \textbf{0.425} \\
        & 0.7\hphantom{$^{\ast}$} & 28.0 $\pm$ 0.7 & 0.416 \\
        & 0.6\hphantom{$^{\ast}$} & 28.8 $\pm$ 0.7 & 0.420 \\
    \bottomrule
    \end{tabular}
    \caption{Machine translation results on EN--DE \texttt{newstest2016} set~\cite{Bojar2016FindingsOT} with larger vocabularies: 8192 for each language separately, and joint vocabularies of sizes 16384, 32768, and 65536 for both languages. For every threshold~$\mathcal{T}$, we report BLEU~\citep{papineni-etal-2002-bleu} and COMET~\citep{rei-etal-2020-comet} scores on the translation task. The best scores are highlighted in \textbf{bold}. Other scores that are not statistically significantly different from the best are also highlighted in \textbf{bold}. If neither of the scores is significantly better, nothing is highlighted. $^{\ast}\mathcal{T} = 1.0$ represents the baseline vanilla BPE without intermediate token removal.}
    \label{tab:en-de-large}
\end{table}

\paragraph{Larger vocabularies.} We also tested Picky BPE with larger vocabularies for the EN--DE task. We used two settings: separate vocabularies for input and output, and joint vocabularies. In both cases, we used total vocabulary sizes 16384, 32768, and 65536. The results for all these experiments are presented in Table~\ref{tab:en-de-large}. As with the smaller vocabulary setting, we see models based on Picky BPE tokenization performing on par with the ones based on the vanilla BPE. In most experiments, our method brings downstream improvements judging by the values of the COMET metric. We also observe by the BLEU scores that for the largest vocabularies of sizes 32768 + 32768 and 65536 the performance is generally worse than with the smaller vocabularies, regardless of the tokenization method. This is likely due to the volume of training data being insufficient for such a large vocabulary. However, in this setting Picky BPE still outperforms vanilla BPE by COMET.

\begin{figure*}[t!]
    \begin{subfigure}[b]{0.48\textwidth}
        \includegraphics[width=0.93\textwidth]{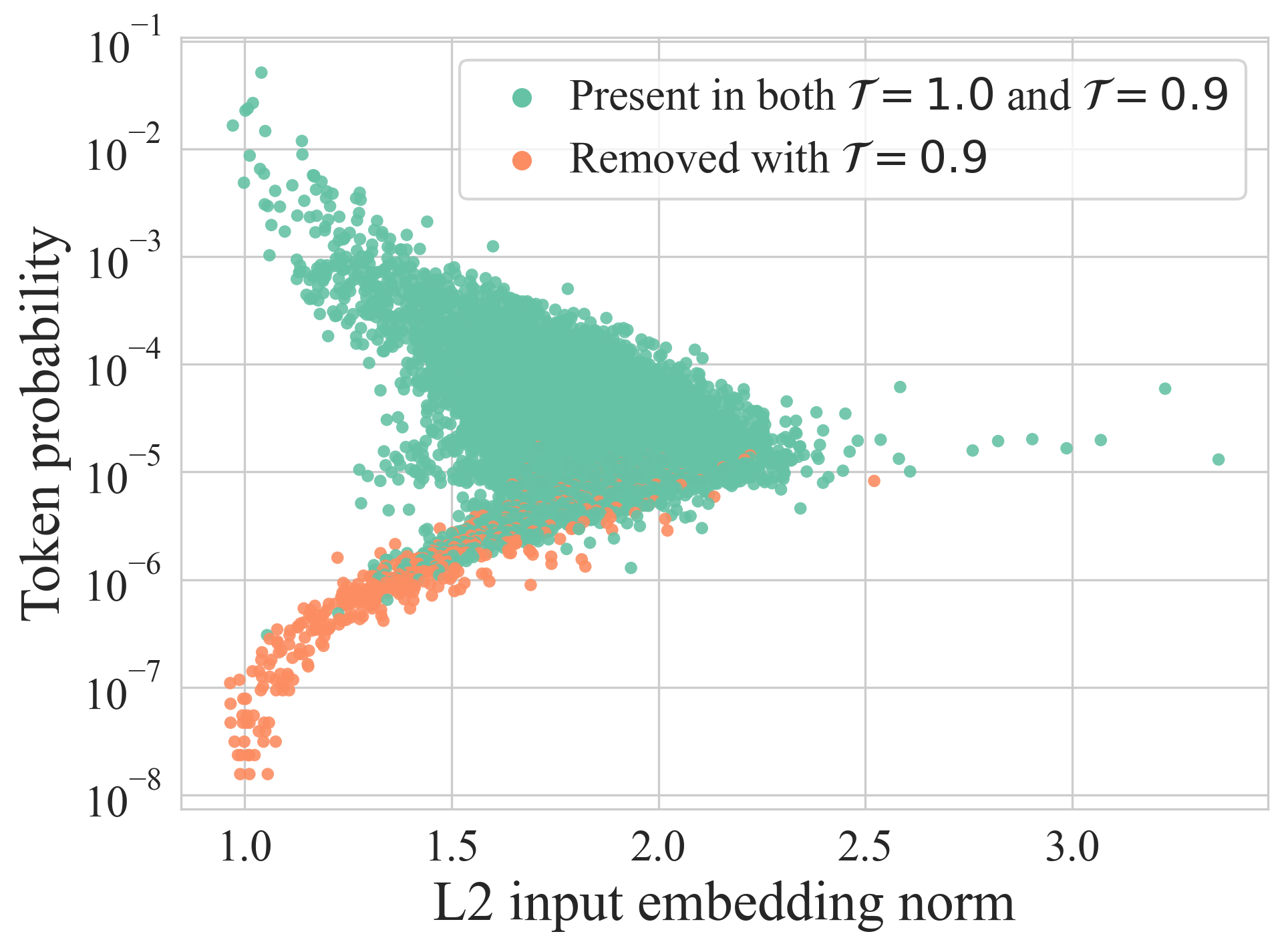}
        \subcaption{Picky BPE tokens when $\mathcal{T} = 1.0$. The tokens that are present when $\mathcal{T} = 1.0$ but removed when $\mathcal{T} = 0.9$ (orange) are generally infrequent and have low L2 embedding norms, thus the majority of them are likely to be undertrained~\citep{land2024fishing}.}
        \label{fig:removed-0-9}
    \end{subfigure}
    \hfill
    \begin{subfigure}[b]{0.48\textwidth}
        \includegraphics[width=0.93\textwidth]{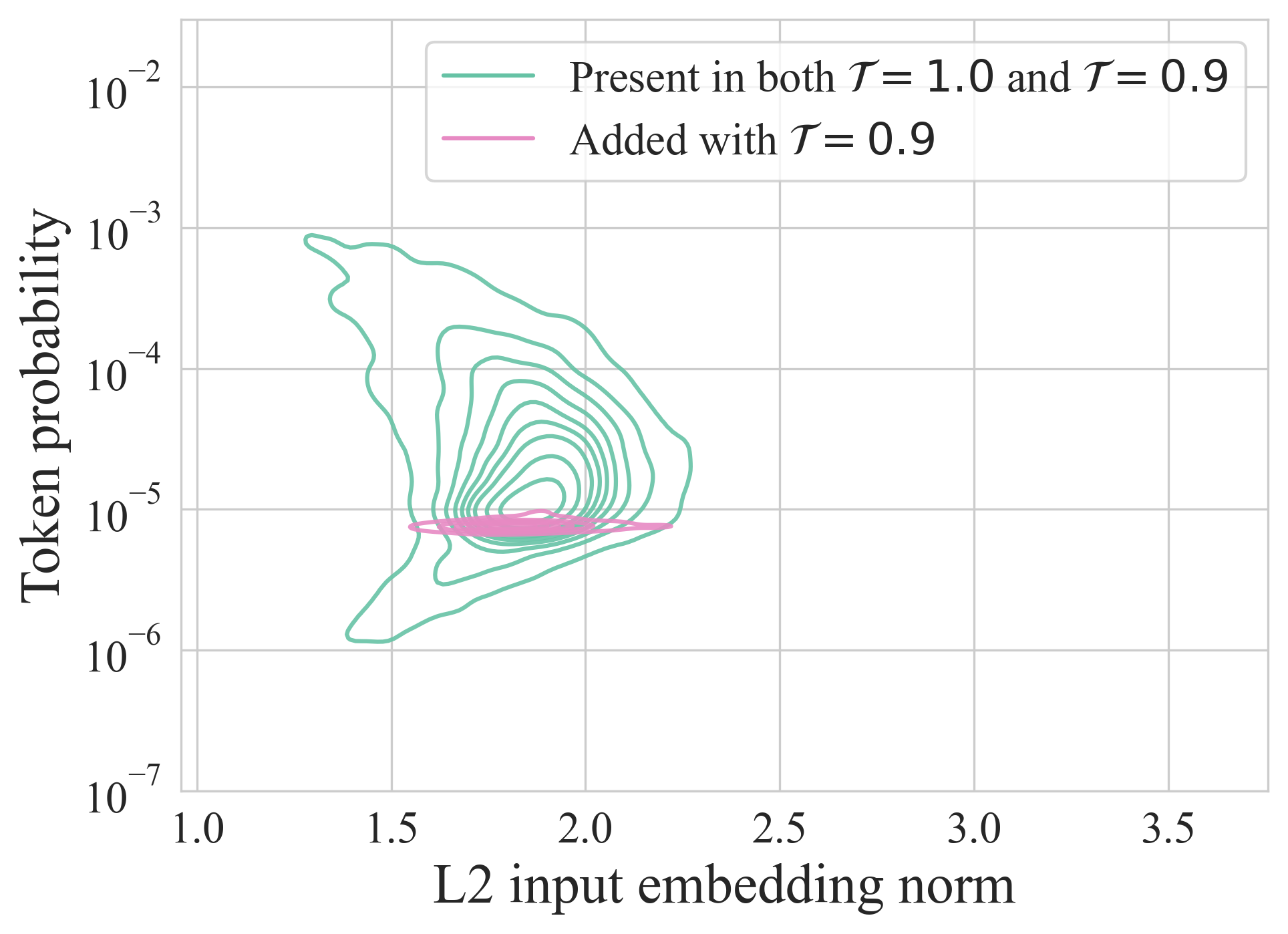}
        \subcaption{Picky BPE tokens when $\mathcal{T} = 0.9$. The tokens that are present when $\mathcal{T} = 0.9$ but not when $\mathcal{T} = 1.0$ (pink) have frequencies and L2-norms of the embeddings close to the blob center and thus are less likely to be under-trained~\citep{land2024fishing}.}
        \label{fig:added-0-9}
    \end{subfigure}
    \caption{Input embedding vectors for Picky BPE tokens with \textbf{(a)} $\mathcal{T} = 1$ and \textbf{(b)} $\mathcal{T} = 0.9$ for English vocabularies of size 16384 in EN--DE experiments with separate vocabularies. For each token we compute its probability in the training corpus (y-axis), and the L2 norm of its embedding vector in the trained model (x-axis).}
    \label{fig:scatter}
\end{figure*}

\begin{figure}[t!]
    \includegraphics[width=0.95\linewidth]{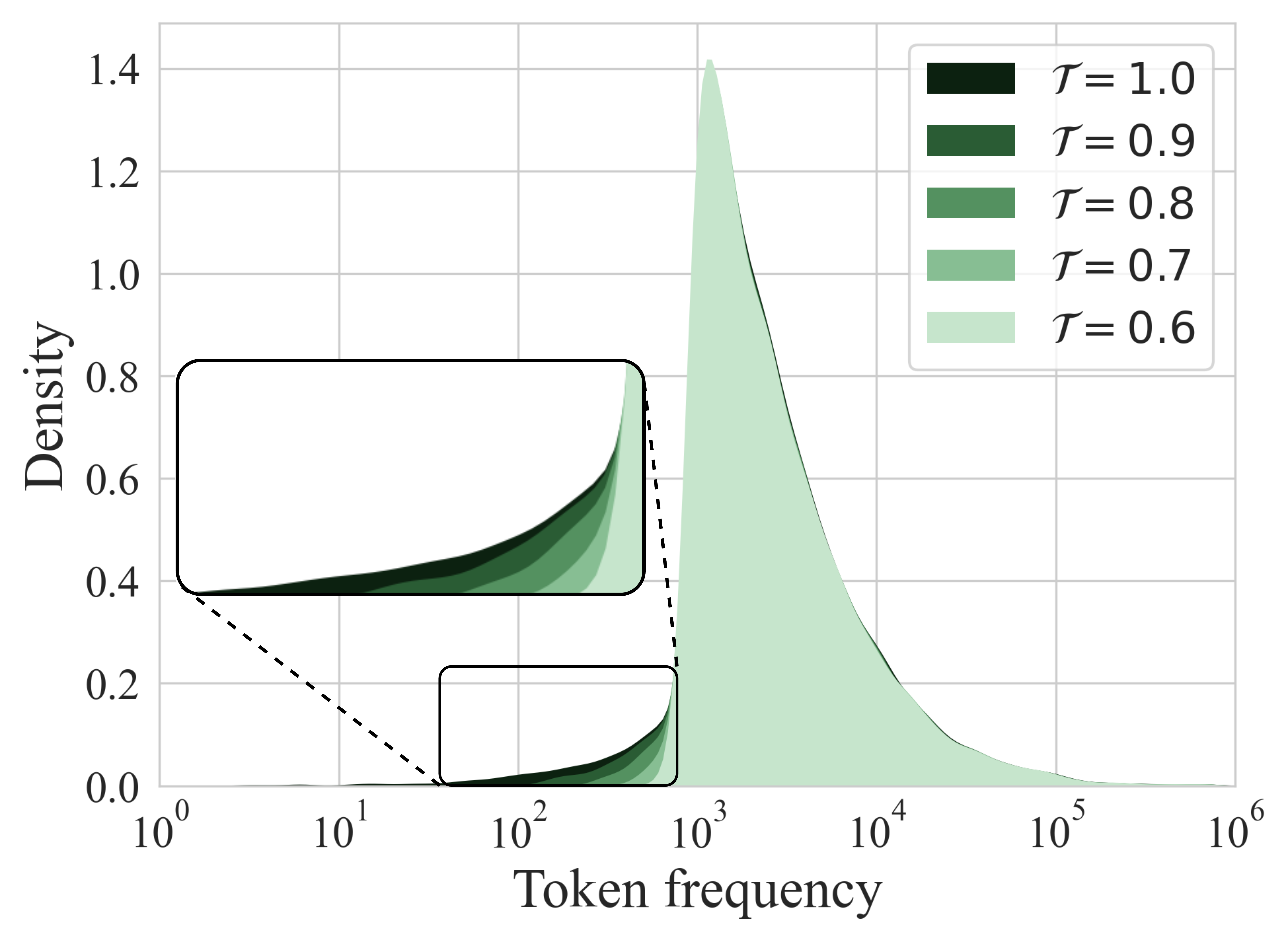}
    \caption{Token frequency distributions for English vocabularies of size 16384 in EN--DE experiments with separate vocabularies for input and output. The left tail becomes less heavy as we decrease the threshold.}
    \label{fig:pdf}
\end{figure}

\section{Under-Trained Tokens} \label{sec:under_trained}

We also test whether Picky BPE decreases the number tokens likely to be under-trained. These tokens can be identified by looking for very low L2 norm of the token embeddings  \citep{land2024fishing}. We plot L2 norms for $\mathcal{T} = 0.9$ in Figure~\ref{fig:scatter} and those for the remaining thresholds in Appendix~\ref{app:under-trained}. There are two groups of low-L2 norm tokens: the first is the low-frequency tokens, which can be seen in the lower left of Figure \ref{fig:removed-0-9}. According to \citet{land2024fishing}, these are the candidates for under-training. There is also a group of the highest-frequency tokens with low L2 norms (top left, Figure \ref{fig:removed-0-9}). We posit that these are general-purpose tokens that occur in a wide variety of contexts, and thus their representations are less specific. It has long been observed that high-frequency words are more likely to have more senses, \textit{i.e.}, meanings \citep{zipf1945meaning}, and thus be more general-purpose.

A large portion of the tokens removed by Picky BPE (Figure~\ref{fig:removed-0-9}) are likely to become under-trained. By contrast, the added tokens (Figure \ref{fig:added-0-9}) have higher L2 norms and higher probability. The high-frequency general tokens are not removed by Picky BPE. 
We argue that Picky BPE reduces the likelihood of under-trained tokens and the risks that come with them, such as increased hallucinations.

We also find that as we lower the threshold for Picky BPE, there is a decrease in the left tail of the token frequency distribution, which represents the low-frequency tokens (Figure~\ref{fig:pdf}). Trimming methods that involve an absolute frequency cut-off, such as the one used by \citet{cognetta2024analysis} and originally proposed in \citet{sennrich-etal-2017-university}, would completely eliminate the left tail and leave an abrupt fall-off on the distribution. We observe that Picky BPE preserves the overall distribution and does not eliminate the left tail. This shows that Picky BPE is not another implementation of the post-training trimming of low-frequency tokens.  

\begin{table}[t!]
    \centering
    \begin{tabular}{ccc}
    \toprule
        & \multirow{3}{*}{
        \begin{tabular}{c}
            \# unique tokens\\
            vs vanilla BPE
        \end{tabular}
        } & \# unique tokens \\
        $\mathcal{T}$ & & vs vanilla BPE \\
        & & + post-trimming \\
        \midrule
        0.9 & 168 (2.1\%)\hphantom{1} & 115 (1.4\%) \\
        0.8 & 391 (4.8\%)\hphantom{1} & 248 (3.0\%) \\
        0.7 & 625 (7.6\%)\hphantom{1} & 393 (4.8\%) \\
        0.6 & 869 (10.6\%) & 588 (7.2\%) \\
    \bottomrule
    \end{tabular}
    \caption{Comparison of tokens from picky BPE and vanilla BPE for joint EN--DE vocabularies of size 8192. For each threshold $\mathcal{T}$, we report the number of unique tokens in the Picky BPE vocabulary compared to the vanilla BPE $(\mathcal{T} = 1)$ with and without low-frequency token trimming on post-processing.}
    \label{tab:ablation}
\end{table}

\begin{table*}[!t]
\centering
\begin{tabular}{@{}cccccccc@{}}
\toprule
\multirow{2.4}{*}{Threshold} & \multirow{2.4}{*}{\# removed} & \multicolumn{2}{c}{Compression $(\downarrow)$} & \multicolumn{3}{c}{\% Word-Initial Tokens} & \multirow{2.4}{*}{\shortstack{Mean Token \\ Length $(\uparrow)$}} \\ \cmidrule(lr){3-4} \cmidrule(lr){5-7}
 && German & English & Dropped $(\downarrow)$ & Added $(\uparrow)$ & Overall $(\uparrow)$ & \\ 
 \midrule
1.0$^{\ast}$ & 0 & 1.000 & 1.000 & --- & --- & 61.5 & 5.38 \\
 \midrule
0.9\hphantom{$^{\ast}$} & 160 & 0.997 & 0.996 & 43.8 & 65.5 & 61.9 & 5.40 \\
0.8\hphantom{$^{\ast}$} & 358 & 0.995 & 0.993 & \textbf{41.1} & \textbf{67.5} & 62.7 & 5.44 \\
0.7\hphantom{$^{\ast}$} & 588 & 0.994 & 0.991 & 42.0 & 66.9 & 63.3 & 5.47 \\
0.6\hphantom{$^{\ast}$} & 805 & \textbf{0.992} & \textbf{0.989} & 42.1 & 64.2 & \textbf{63.6} & \textbf{5.50} \\ \bottomrule
\end{tabular}
\caption{Token quality evaluation on EN--DE tokenizers with joint vocabularies of size 8192. Compression scores are reported as corpus token counts of the \texttt{newstest2016} set relative to the vanilla BPE, such that 1 indicates the same compression rate. We report the proportion of word-initial tokens out of dropped tokens, added tokens, and out of the whole vocabulary along with the mean token length in characters. $^{\ast}\mathcal{T} = 1.0$ represents the baseline vanilla BPE without intermediate token removal.
}
\label{tab:qualities}
\end{table*}

Table \ref{tab:ablation} shows the difference between Picky BPE and vanilla BPE with and without post-processing trimming. By post-trimming we mean training the vanilla BPE to have a larger vocabulary with further trimming low-frequency tokens to achieve the desired vocabulary size. We train the initial tokenizer so the number of additional tokens is the number of replaced tokens from the corresponding Picky BPE tokenizer. Through both differences in the number of replaced tokens in the two different strategies, we show that Picky BPE is not simply a different implementation of the post-trimming akin to~\citet{cognetta2024analysis}, but it leads to a fundamentally different resulting vocabulary. 

\section{Features of Picky BPE} \label{sec:features}

\paragraph{Text Compression.}

Text compression is generally considered to be an important aspect of tokenizer evaluation \citep{galle-2019-investigating, goldman2024unpacking}, and language models that compress more have been shown to have better performance \citep{liang-etal-2023-xlm, goldman2024unpacking}. We use \textit{corpus token count} (CTC; \citet{schmidt2024tokenization}) to measure compression. CTC, also called sequence length, is the number of tokens needed to represent a given text. The fewer tokens are needed, the better the compression.

Table \ref{tab:qualities} shows the changes in compression as a percentage relative to the tokenizer of the same vocabulary size with a threshold of 1, all for EN--DE vocabularies of size 8192. We report additional compression rates in Appendix~\ref{app:compression}. We find that Picky BPE shows no loss in compression. This is an improvement over the method in \citet{cognetta2024analysis}, which shows worse compression after vocabulary trimming. 

\paragraph{Token Qualities.}

In addition to the above metrics, we compare the tokens themselves. One quality of interest is the proportion of word-initial tokens, which are stored in the tokenizer with an underscore at the beginning to represent a space character. \citet{yehezkel-pinter-2023-incorporating} also notice that their trimming procedure leads to an increased number of word-initial tokens. 

In Table \ref{tab:qualities}, we also report the percentage of word-initial tokens from the added and removed tokens as well as overall proportions for the EN--DE vocabulary of size 8192. We report results for the other experiments in Appendix \ref{app:word-initial}. We find that dropped tokens are far less likely to be word-initial than added tokens. Therefore, Picky BPE is adding more word-initial tokens than it is removing. As the threshold is lowered, we see slightly fewer word-initial tokens added to the vocabulary. This might be due to the intensive removals happening with lower thresholds. In the overall rates of word-initial tokens, we see a slight increase as $\mathcal{T}$ goes down. 

Upon inspection of the added tokens, we see that many of the word-initial tokens are also complete, meaningful words, for example \texttt{\textunderscore renovated}, \texttt{\textunderscore overcoat}, \texttt{\textunderscore cognition}, and \texttt{\textunderscore unconventional}. Increased rates of word-initial tokens may be indicative of improved token quality. 

We also found that many of the tokens removed by Picky BPE were intermediate, much like \texttt{entucky} (Figure \ref{fig:kentucky}). These tokens are relatively long and only occur in the context of a longer token that is also present in the vocabulary. Often, these tokens are missing only one or two characters relative to the full word. We find word-initial and word-medial intermediate tokens, \textit{e.g.}, \texttt{\textunderscore Chicag}, \texttt{\textunderscore algorith}, \texttt{roprietary}, \texttt{omenclature} (cf. `Chicago', `algorithm', `proprietary', `nomenclature').

Following \citet{bostrom-durrett-2020-byte}, we also measure mean token length. They argue that longer mean token length is associated with gold-standard morphologically-aligned tokenization, and thus with better token quality. Additionally, longer tokens on average will lead to increased compression, as a text of a fixed length can be represented with fewer, longer tokens. We find that the mean token length slightly but consistently increases as we lower the threshold (see Table~\ref{tab:qualities}).
We report additional mean token length results in Appendix~\ref{app:token_len}. 

We additionally compare Picky BPE with Unigram tokenization in Table~\ref{tab:sp}. Unigram tokenization seems to have longer tokens with a higher proportion of word-initial tokens. However, it drastically worsens the compression. We hypothesize that Unigram adds many meaningful full-word tokens which are not optimal for the text compression under the restriction of the vocabulary size.

\begin{table}[t!]
    \centering
    \begin{tabular}{@{}lcccc@{}}
    \toprule
    \multirow{2.4}{*}{Method} & \multicolumn{2}{c}{CTC ($\downarrow$)} & \multirow{2.4}{*}{
        \shortstack{
            \% Word-\\
            initial ($\uparrow$)
        }} & \multirow{2.4}{*}{
        \shortstack{
            Mean\\
            len ($\uparrow$)
        }}\\ 
    \cmidrule(lr){2-3}
     & EN & DE & & \\ 
    \midrule
    Unigram & 1.143 & 1.124 & \textbf{75.6} & \textbf{7.73} \\
     \midrule
    $\mathcal{T} = 1.0$ & 1.000 & 1.000 & 72.2 & 6.85 \\
     \midrule
    $\mathcal{T} = 0.9$ & 0.997 & 0.998 & 72.8 & 6.88 \\
    $\mathcal{T} = 0.8$ & 0.996 & 0.998 & 73.2 & 6.91 \\
    $\mathcal{T} = 0.7$ & 0.994 & 0.997 & 73.6 & 6.94 \\
    $\mathcal{T} = 0.6$ & \textbf{0.992} & \textbf{0.996} & 73.9 & 6.95 \\ \bottomrule
    \end{tabular}
    \caption{Comparing Picky BPE and Unigram~\cite{kudo-2018-subword} on joint EN--DE vocabularies of size 32768. We report corpus token counts (CTC) on the \texttt{newstest2016} set relative to the vanilla BPE $(\mathcal{T} = 1.0)$, percentage of word-initial tokens, and mean token length (``Mean len'' in the Table).}
    \label{tab:sp}
\end{table}

\section{Discussion}
\label{sec:discussion}

We believe Picky BPE would be beneficial for Large Language Models (LLMs), however, the lack of computational resources does not allow us to carry out a side-by-side comparison. Instead, we provide a series of experiments that we believe illustrate key properties of the proposed method. To put these results into perspective, we want to reiterate two core aspects of the provided experiments: first, there is no universal methodology that could assess tokenizer quality; second, the inefficiencies associated with undertrained tokens discussed by \citet{land2024fishing} depend on the size of vocabulary relative to the size of training data.

\paragraph{Evaluating tokenizers.}

It is not always clear how to best compare different tokenizers \citep{zouhar-etal-2023-tokenization}. One approach is training models for each tokenizer and evaluating downstream performance, \textit{e.g.}, \citet{goldman2024unpacking}. However, these results may be driven by confounding factors, such as differences in compression leading to the model effectively being trained on less text \citep{petrov2024language}, and downstream task results may also be task-specific. The second general approach to evaluating tokenizers is to evaluate some quality of the tokenizer's output such as fertility (average number of tokens per word; \citet{rust-etal-2021-good}), similarity of tokenizer boundaries to morphological boundaries \citep{hofmann-etal-2021-superbizarre}, and cognitive plausibility of tokens \citep{beinborn-pinter-2023-analyzing}. There is no consensus about which metric(s) provide the best overall estimation of tokenizer quality.

\paragraph{Role of undertrained tokens.} We achieved better or equal performance on machine translation with small vocabularies compared to the vanilla BPE. However, we did not improve the performance with a large vocabulary. The restriction on vocabulary size was set intentionally to reduce redundancy and ensure all tokens receive enough training. We expect to see the same effect with LLMs as their large vocabulary size corresponds to the massive scale of training data and model size. This is well-justified by our analysis of under-trained tokens in response to the exploration of LLMs by~\citet{land2024fishing}. We also witness improved token quality that comes with our method, which does not affect text compression, see comparison to the Unigram tokenization in \S\ref{sec:features}.

\section{Conclusion}

In this paper, we propose a novel tokenization algorithm, Picky BPE, which refines vocabulary during tokenizer training targeting intermediate tokens. Our results show that our algorithm may improve downstream performance in a setting of limited vocabulary, which we can extrapolate on larger vocabularies given enough training. Our method also mitigates the issue of under-trained tokens, efficiently removing them during training, and improves token quality and text compression, filling the freed vocabulary space with meaningful tokens with higher frequency. These factors suggest that Picky BPE can be considered for larger models to improve downstream performance and safety and avoid undesired behavior, \textit{e.g.}, hallucinations.

\section{Limitations}
Picky BPE behavior depends on the choice of threshold $\mathcal{T}$. Even though the threshold is relative and mostly intuitive in use, one must consider that with lower thresholds the probability of eliminating useful tokens grows and the behavior becomes less stable. Therefore, it is important to start with safer larger thresholds, analyzing the tokenization using vocabulary-related measures.

In this paper, the only downstream task we evaluate our models on is translation. Training a larger language model and evaluating it on other downstream tasks may show different patterns. This may allow us to better understand the contribution of Picky BPE as well as its potential drawbacks. 

\citet{rust-etal-2021-good} show that different tasks have variable correlation with tokenizer evaluations like fertility. To the best of our knowledge, there is not enough empirical work to determine which tasks would be most informative for evaluating tokenizer quality. This is an important area for future work. 

Our experiments are also limited to a relatively small set of languages. We selected pairs of languages that were typologically varied and used different writing systems, however, all the languages are spoken in Europe. Future work should evaluate whether a larger and more diverse sample of languages exhibit the same trends as in this paper. 

\section*{Acknowledgments}

The authors acknowledge the HPC resource allocation by Erlangen National High-Performance Computing Center (NHR@FAU) of the Friedrich-Alexander-Universität Erlangen-Nürnberg (FAU) (joint project with the Center for Artificial Intelligence (CAIRO), THWS) and Jean Zay (compute grant \#GC011015451).
The authors would also like to thank the other members of the PleIAs Research team for helpful discussion and feedback. 

\bibliography{custom}

\begin{thebibliography}{49}
\providecommand{\natexlab}[1]{#1}

\bibitem[{Ali et~al.(2023)Ali, Fromm, Thellmann, Rutmann, L{\"u}bbering, Leveling, Klug, Ebert, Doll, Buschhoff et~al.}]{ali2023tokenizer}
Mehdi Ali, Michael Fromm, Klaudia Thellmann, Richard Rutmann, Max L{\"u}bbering, Johannes Leveling, Katrin Klug, Jan Ebert, Niclas Doll, Jasper~Schulze Buschhoff, et~al. 2023.
\newblock \href {https://arxiv.org/pdf/2310.08754} {Tokenizer choice for {{LLM}} training: Negligible or {C}rucial?}
\newblock \emph{arXiv preprint arXiv:2310.08754}.

\bibitem[{Bauwens and Delobelle(2024)}]{bauwens-delobelle-2024-bpe}
Thomas Bauwens and Pieter Delobelle. 2024.
\newblock \href {https://doi.org/10.18653/v1/2024.naacl-long.324} {{BPE}-knockout: Pruning pre-existing {BPE} tokenisers with backwards-compatible morphological semi-supervision}.
\newblock In \emph{Proceedings of the 2024 Conference of the North American Chapter of the Association for Computational Linguistics: Human Language Technologies (Volume 1: Long Papers)}, pages 5810--5832, Mexico City, Mexico. Association for Computational Linguistics.

\bibitem[{Beinborn and Pinter(2023)}]{beinborn-pinter-2023-analyzing}
Lisa Beinborn and Yuval Pinter. 2023.
\newblock \href {https://doi.org/10.18653/v1/2023.emnlp-main.272} {Analyzing cognitive plausibility of subword tokenization}.
\newblock In \emph{Proceedings of the 2023 Conference on Empirical Methods in Natural Language Processing}, pages 4478--4486, Singapore. Association for Computational Linguistics.

\bibitem[{Bojar et~al.(2016)Bojar, Chatterjee, Federmann, Graham, Haddow, Huck, Jimeno-Yepes, Koehn, Logacheva, Monz, Negri, N{\'e}v{\'e}ol, Neves, Popel, Post, Rubino, Scarton, Specia, Turchi, Verspoor, and Zampieri}]{Bojar2016FindingsOT}
Ondrej Bojar, Rajen Chatterjee, Christian Federmann, Yvette Graham, Barry Haddow, Matthias Huck, Antonio Jimeno-Yepes, Philipp Koehn, Varvara Logacheva, Christof Monz, Matteo Negri, Aur{\'e}lie N{\'e}v{\'e}ol, Mariana Neves, Martin Popel, Matt Post, Rapha{\"e}l Rubino, Carolina Scarton, Lucia Specia, Marco Turchi, Karin~M. Verspoor, and Marcos Zampieri. 2016.
\newblock \href {https://api.semanticscholar.org/CorpusID:14421595} {Findings of the 2016 {C}onference on {M}achine {T}ranslation}.
\newblock In \emph{Conference on Machine Translation}.

\bibitem[{Bostrom and Durrett(2020)}]{bostrom-durrett-2020-byte}
Kaj Bostrom and Greg Durrett. 2020.
\newblock \href {https://doi.org/10.18653/v1/2020.findings-emnlp.414} {Byte {P}air {E}ncoding is {S}uboptimal for {L}anguage {M}odel {P}retraining}.
\newblock In \emph{Findings of the Association for Computational Linguistics: EMNLP 2020}, pages 4617--4624, Online. Association for Computational Linguistics.

\bibitem[{Cognetta et~al.(2024)Cognetta, Hiraoka, Okazaki, Sennrich, and Pinter}]{cognetta2024analysis}
Marco Cognetta, Tatsuya Hiraoka, Naoaki Okazaki, Rico Sennrich, and Yuval Pinter. 2024.
\newblock \href {https://arxiv.org/pdf/2404.00397} {An {A}nalysis of {BPE} {V}ocabulary {T}rimming in {N}eural {M}achine {T}ranslation}.
\newblock \emph{arXiv preprint arXiv:2404.00397}.

\bibitem[{Gage(1994)}]{gage1994new}
Philip Gage. 1994.
\newblock A new algorithm for data compression.
\newblock \emph{The C Users Journal}, 12(2):23--38.

\bibitem[{Gall{\'e}(2019)}]{galle-2019-investigating}
Matthias Gall{\'e}. 2019.
\newblock \href {https://doi.org/10.18653/v1/D19-1141} {{I}nvestigating the {E}ffectiveness of {BPE}: {T}he {P}ower of {S}horter {S}equences}.
\newblock In \emph{Proceedings of the 2019 Conference on Empirical Methods in Natural Language Processing and the 9th International Joint Conference on Natural Language Processing (EMNLP-IJCNLP)}, pages 1375--1381, Hong Kong, China. Association for Computational Linguistics.

\bibitem[{Geiping et~al.(2024)Geiping, Stein, Shu, Saifullah, Wen, and Goldstein}]{geiping2024coercing}
Jonas Geiping, Alex Stein, Manli Shu, Khalid Saifullah, Yuxin Wen, and Tom Goldstein. 2024.
\newblock \href {https://arxiv.org/pdf/2402.14020} {Coercing {LLMs} to do and reveal (almost) anything}.
\newblock \emph{arXiv preprint arXiv:2402.14020}.

\bibitem[{Goldman et~al.(2024)Goldman, Caciularu, Eyal, Cao, Szpektor, and Tsarfaty}]{goldman2024unpacking}
Omer Goldman, Avi Caciularu, Matan Eyal, Kris Cao, Idan Szpektor, and Reut Tsarfaty. 2024.
\newblock \href {https://arxiv.org/pdf/2403.06265} {Unpacking {T}okenization: {E}valuating {T}ext {C}ompression and its {C}orrelation with {M}odel {P}erformance}.
\newblock \emph{arXiv preprint arXiv:2403.06265}.

\bibitem[{Goyal et~al.(2022)Goyal, Gao, Chaudhary, Chen, Wenzek, Ju, Krishnan, Ranzato, Guzmán, and Fan}]{10.1162/tacl_a_00474}
Naman Goyal, Cynthia Gao, Vishrav Chaudhary, Peng-Jen Chen, Guillaume Wenzek, Da~Ju, Sanjana Krishnan, Marc’Aurelio Ranzato, Francisco Guzmán, and Angela Fan. 2022.
\newblock \href {https://doi.org/10.1162/tacl_a_00474} {{The Flores-101 Evaluation Benchmark for Low-Resource and Multilingual Machine Translation}}.
\newblock \emph{Transactions of the Association for Computational Linguistics}, 10:522--538.

\bibitem[{Groeneveld et~al.(2024)Groeneveld, Beltagy, Walsh, Bhagia, Kinney, Tafjord, Jha, Ivison, Magnusson, Wang et~al.}]{groeneveld2024olmo}
Dirk Groeneveld, Iz~Beltagy, Pete Walsh, Akshita Bhagia, Rodney Kinney, Oyvind Tafjord, Ananya~Harsh Jha, Hamish Ivison, Ian Magnusson, Yizhong Wang, et~al. 2024.
\newblock \href {https://arxiv.org/pdf/2402.00838} {{OLM}o: {A}ccelerating the {S}cience of {L}anguage {M}odels}.
\newblock \emph{arXiv preprint arXiv:2402.00838}.

\bibitem[{Hofmann et~al.(2021)Hofmann, Pierrehumbert, and Sch{\"u}tze}]{hofmann-etal-2021-superbizarre}
Valentin Hofmann, Janet Pierrehumbert, and Hinrich Sch{\"u}tze. 2021.
\newblock \href {https://doi.org/10.18653/v1/2021.acl-long.279} {Superbizarre is not superb: Derivational morphology improves {BERT}{'}s interpretation of complex words}.
\newblock In \emph{Proceedings of the 59th Annual Meeting of the Association for Computational Linguistics and the 11th International Joint Conference on Natural Language Processing (Volume 1: Long Papers)}, pages 3594--3608, Online. Association for Computational Linguistics.

\bibitem[{Hofmann et~al.(2022)Hofmann, Schuetze, and Pierrehumbert}]{hofmann2022embarrassingly}
Valentin Hofmann, Hinrich Schuetze, and Janet Pierrehumbert. 2022.
\newblock \href {https://doi.org/10.18653/v1/2022.acl-short.43} {An {E}mbarrassingly {S}imple {M}ethod to {M}itigate {U}ndesirable {P}roperties of {P}retrained {L}anguage {M}odel {T}okenizers}.
\newblock In \emph{Proceedings of the 60th Annual Meeting of the Association for Computational Linguistics (Volume 2: Short Papers)}, pages 385--393, Dublin, Ireland. Association for Computational Linguistics.

\bibitem[{Koehn(2004)}]{koehn-2004-statistical}
Philipp Koehn. 2004.
\newblock \href {https://aclanthology.org/W04-3250} {Statistical {S}ignificance {T}ests for {M}achine {T}ranslation {E}valuation}.
\newblock In \emph{Proceedings of the 2004 Conference on Empirical Methods in Natural Language Processing}, pages 388--395, Barcelona, Spain. Association for Computational Linguistics.

\bibitem[{Korotkova and Fishel(2024)}]{estonian_centric_mt}
Elizaveta Korotkova and Mark Fishel. 2024.
\newblock \href {https://eamt2024.github.io/proceedings/vol1.pdf#page=669} {Estonian-{C}entric {M}achine {T}ranslation: {D}ata, {M}odels, and {C}hallenges}.
\newblock In \emph{Proceedings of the 25th Annual Conference of the European Association for Machine Translation (Volume 1: Research And Implementations \& Case Studies)}, pages 647--660, Sheffield, UK. European Association for Machine Translation.

\bibitem[{Kudo(2018)}]{kudo-2018-subword}
Taku Kudo. 2018.
\newblock \href {https://doi.org/10.18653/v1/P18-1007} {Subword {R}egularization: {I}mproving {N}eural network {T}ranslation {M}odels with {M}ultiple {S}ubword {C}andidates}.
\newblock In \emph{Proceedings of the 56th Annual Meeting of the Association for Computational Linguistics (Volume 1: Long Papers)}, pages 66--75, Melbourne, Australia. Association for Computational Linguistics.

\bibitem[{Kudo and Richardson(2018)}]{kudo-richardson-2018-sentencepiece}
Taku Kudo and John Richardson. 2018.
\newblock \href {https://doi.org/10.18653/v1/D18-2012} {{S}entence{P}iece: A simple and language independent subword tokenizer and detokenizer for {N}eural {T}ext {P}rocessing}.
\newblock In \emph{Proceedings of the 2018 Conference on Empirical Methods in Natural Language Processing: System Demonstrations}, pages 66--71, Brussels, Belgium. Association for Computational Linguistics.

\bibitem[{Land and Bartolo(2024)}]{land2024fishing}
Sander Land and Max Bartolo. 2024.
\newblock \href {https://arxiv.org/pdf/2405.05417} {Fishing for {M}agikarp: {A}utomatically {D}etecting {U}nder-trained {T}okens in {L}arge {L}anguage {M}odels}.
\newblock \emph{arXiv preprint arXiv:2405.05417}.

\bibitem[{Li et~al.(2024)Li, Liu, Deng, Zhang, Song, Shi, Wang, Li, Liu, and Wang}]{li2024glitch}
Yuxi Li, Yi~Liu, Gelei Deng, Ying Zhang, Wenjia Song, Ling Shi, Kailong Wang, Yuekang Li, Yang Liu, and Haoyu Wang. 2024.
\newblock \href {https://doi.org/10.1145/3660799} {Glitch tokens in large language models: Categorization taxonomy and effective detection}.
\newblock \emph{Proc. ACM Softw. Eng.}, 1(FSE).

\bibitem[{Lian et~al.(2024)Lian, Xiong, Niu, Mo, Su, Lin, Liu, Chen, and Ding}]{lian2024scaffold}
Haoran Lian, Yizhe Xiong, Jianwei Niu, Shasha Mo, Zhenpeng Su, Zijia Lin, Peng Liu, Hui Chen, and Guiguang Ding. 2024.
\newblock Scaffold-bpe: Enhancing byte pair encoding with simple and effective scaffold token removal.
\newblock \emph{arXiv preprint arXiv:2404.17808}.

\bibitem[{Liang et~al.(2023)Liang, Gonen, Mao, Hou, Goyal, Ghazvininejad, Zettlemoyer, and Khabsa}]{liang-etal-2023-xlm}
Davis Liang, Hila Gonen, Yuning Mao, Rui Hou, Naman Goyal, Marjan Ghazvininejad, Luke Zettlemoyer, and Madian Khabsa. 2023.
\newblock \href {https://doi.org/10.18653/v1/2023.emnlp-main.813} {{XLM}-{V}: {O}vercoming the {V}ocabulary {B}ottleneck in {M}ultilingual {M}asked {L}anguage {M}odels}.
\newblock In \emph{Proceedings of the 2023 Conference on Empirical Methods in Natural Language Processing}, pages 13142--13152, Singapore. Association for Computational Linguistics.

\bibitem[{Ott et~al.(2019)Ott, Edunov, Baevski, Fan, Gross, Ng, Grangier, and Auli}]{ott-etal-2019-fairseq}
Myle Ott, Sergey Edunov, Alexei Baevski, Angela Fan, Sam Gross, Nathan Ng, David Grangier, and Michael Auli. 2019.
\newblock \href {https://doi.org/10.18653/v1/N19-4009} {fairseq: {A} {F}ast, {E}xtensible {T}oolkit for {S}equence {M}odeling}.
\newblock In \emph{Proceedings of the 2019 Conference of the North {A}merican Chapter of the Association for Computational Linguistics (Demonstrations)}, pages 48--53, Minneapolis, Minnesota. Association for Computational Linguistics.

\bibitem[{Pang and Vuli{\'c}(2024)}]{pang2024specialising}
Jiayun Pang and Ivan Vuli{\'c}. 2024.
\newblock \href {https://arxiv.org/pdf/2405.10625} {Specialising and {A}nalysing {I}nstruction-{T}uned and {B}yte-{L}evel {L}anguage {M}odels for {O}rganic {R}eaction {P}rediction}.
\newblock \emph{arXiv preprint arXiv:2405.10625}.

\bibitem[{Papineni et~al.(2002)Papineni, Roukos, Ward, and Zhu}]{papineni-etal-2002-bleu}
Kishore Papineni, Salim Roukos, Todd Ward, and Wei-Jing Zhu. 2002.
\newblock \href {https://doi.org/10.3115/1073083.1073135} {{B}leu: a method for automatic evaluation of machine translation}.
\newblock In \emph{Proceedings of the 40th Annual Meeting of the Association for Computational Linguistics}, pages 311--318, Philadelphia, Pennsylvania, USA. Association for Computational Linguistics.

\bibitem[{Petrov et~al.(2023)Petrov, La~Malfa, Torr, and Bibi}]{petrov2024language}
Aleksandar Petrov, Emanuele La~Malfa, Philip Torr, and Adel Bibi. 2023.
\newblock \href {https://proceedings.neurips.cc/paper_files/paper/2023/file/74bb24dca8334adce292883b4b651eda-Paper-Conference.pdf} {Language {M}odel {T}okenizers {I}ntroduce {U}nfairness {B}etween {L}anguages}.
\newblock In \emph{Advances in Neural Information Processing Systems}, volume~36, pages 36963--36990. Curran Associates, Inc.

\bibitem[{Post(2018)}]{post-2018-call}
Matt Post. 2018.
\newblock \href {https://doi.org/10.18653/v1/W18-6319} {A call for clarity in reporting {BLEU} scores}.
\newblock In \emph{Proceedings of the Third Conference on Machine Translation: Research Papers}, pages 186--191, Brussels, Belgium. Association for Computational Linguistics.

\bibitem[{Provilkov et~al.(2020)Provilkov, Emelianenko, and Voita}]{provilkov-etal-2020-bpe}
Ivan Provilkov, Dmitrii Emelianenko, and Elena Voita. 2020.
\newblock \href {https://doi.org/10.18653/v1/2020.acl-main.170} {{BPE}-dropout: Simple and effective subword regularization}.
\newblock In \emph{Proceedings of the 58th Annual Meeting of the Association for Computational Linguistics}, pages 1882--1892, Online. Association for Computational Linguistics.

\bibitem[{Rajaraman et~al.(2024)Rajaraman, Jiao, and Ramchandran}]{rajaraman2024toward}
Nived Rajaraman, Jiantao Jiao, and Kannan Ramchandran. 2024.
\newblock \href {https://arxiv.org/pdf/2404.08335} {Toward a {T}heory of {T}okenization in {LLM}s}.
\newblock \emph{arXiv preprint arXiv:2404.08335}.

\bibitem[{Rei et~al.(2020)Rei, Stewart, Farinha, and Lavie}]{rei-etal-2020-comet}
Ricardo Rei, Craig Stewart, Ana~C Farinha, and Alon Lavie. 2020.
\newblock \href {https://doi.org/10.18653/v1/2020.emnlp-main.213} {{COMET}: {A} {N}eural {F}ramework for {MT} evaluation}.
\newblock In \emph{Proceedings of the 2020 Conference on Empirical Methods in Natural Language Processing (EMNLP)}, pages 2685--2702, Online. Association for Computational Linguistics.

\bibitem[{Rumbelow and Watkins(2023)}]{rumbelow2023solidgoldmagikarp}
Jessica Rumbelow and Matthew Watkins. 2023.
\newblock \href {https://www.lesswrong.com/posts/aPeJE8bSo6rAFoLqg/solidgoldmagikarp-plus-prompt-generation} {Solidgoldmagikarp (plus, prompt generation)}.

\bibitem[{Rust et~al.(2021)Rust, Pfeiffer, Vuli{\'c}, Ruder, and Gurevych}]{rust-etal-2021-good}
Phillip Rust, Jonas Pfeiffer, Ivan Vuli{\'c}, Sebastian Ruder, and Iryna Gurevych. 2021.
\newblock \href {https://doi.org/10.18653/v1/2021.acl-long.243} {How {G}ood is {Y}our {T}okenizer? {O}n the {M}onolingual {P}erformance of {M}ultilingual {L}anguage {M}odels}.
\newblock In \emph{Proceedings of the 59th Annual Meeting of the Association for Computational Linguistics and the 11th International Joint Conference on Natural Language Processing (Volume 1: Long Papers)}, pages 3118--3135, Online. Association for Computational Linguistics.

\bibitem[{Schmidt et~al.(2024)Schmidt, Reddy, Zhang, Alameddine, Uzan, Pinter, and Tanner}]{schmidt2024tokenization}
Craig~W Schmidt, Varshini Reddy, Haoran Zhang, Alec Alameddine, Omri Uzan, Yuval Pinter, and Chris Tanner. 2024.
\newblock \href {https://arxiv.org/pdf/2402.18376v1} {Tokenization {I}s {M}ore {T}han {C}ompression}.
\newblock \emph{arXiv preprint arXiv:2402.18376}.

\bibitem[{Sennrich et~al.(2017)Sennrich, Birch, Currey, Germann, Haddow, Heafield, Miceli~Barone, and Williams}]{sennrich-etal-2017-university}
Rico Sennrich, Alexandra Birch, Anna Currey, Ulrich Germann, Barry Haddow, Kenneth Heafield, Antonio~Valerio Miceli~Barone, and Philip Williams. 2017.
\newblock \href {https://doi.org/10.18653/v1/W17-4739} {The {U}niversity of {E}dinburgh{'}s neural {MT} systems for {WMT}17}.
\newblock In \emph{Proceedings of the Second Conference on Machine Translation}, pages 389--399, Copenhagen, Denmark. Association for Computational Linguistics.

\bibitem[{Sennrich et~al.(2016)Sennrich, Haddow, and Birch}]{sennrich-etal-2016-neural}
Rico Sennrich, Barry Haddow, and Alexandra Birch. 2016.
\newblock \href {https://doi.org/10.18653/v1/P16-1162} {Neural {M}achine {T}ranslation of {R}are {W}ords with {S}ubword {U}nits}.
\newblock In \emph{Proceedings of the 54th Annual Meeting of the Association for Computational Linguistics (Volume 1: Long Papers)}, pages 1715--1725, Berlin, Germany. Association for Computational Linguistics.

\bibitem[{Sennrich and Zhang(2019)}]{sennrich-zhang-2019-revisiting}
Rico Sennrich and Biao Zhang. 2019.
\newblock \href {https://doi.org/10.18653/v1/P19-1021} {Revisiting low-resource neural machine translation: A case study}.
\newblock In \emph{Proceedings of the 57th Annual Meeting of the Association for Computational Linguistics}, pages 211--221, Florence, Italy. Association for Computational Linguistics.

\bibitem[{Shao et~al.(2024)Shao, Xiao, Liu, and Zhang}]{shao2024flexibly}
Ninglu Shao, Shitao Xiao, Zheng Liu, and Peitian Zhang. 2024.
\newblock \href {https://arxiv.org/pdf/2401.07793} {Flexibly {S}caling {L}arge {L}anguage {M}odels {C}ontexts {T}hrough {E}xtensible {T}okenization}.
\newblock \emph{arXiv preprint arXiv:2401.07793}.

\bibitem[{Singh and Strouse(2024)}]{singh2024tokenization}
Aaditya~K Singh and DJ~Strouse. 2024.
\newblock \href {https://arxiv.org/pdf/2402.14903} {Tokenization counts: the impact of tokenization on arithmetic in frontier {LLM}s}.
\newblock \emph{arXiv preprint arXiv:2402.14903}.

\bibitem[{Song et~al.(2021)Song, Salcianu, Song, Dopson, and Zhou}]{song-etal-2021-fast}
Xinying Song, Alex Salcianu, Yang Song, Dave Dopson, and Denny Zhou. 2021.
\newblock \href {https://doi.org/10.18653/v1/2021.emnlp-main.160} {Fast {W}ord{P}iece {T}okenization}.
\newblock In \emph{Proceedings of the 2021 Conference on Empirical Methods in Natural Language Processing}, pages 2089--2103, Online and Punta Cana, Dominican Republic. Association for Computational Linguistics.

\bibitem[{Toraman et~al.(2023)Toraman, Yilmaz, \c{S}ahinu\c{c}, and Ozcelik}]{toraman2023impact}
Cagri Toraman, Eyup~Halit Yilmaz, Furkan \c{S}ahinu\c{c}, and Oguzhan Ozcelik. 2023.
\newblock \href {https://doi.org/10.1145/3578707} {Impact of tokenization on language models: An analysis for turkish}.
\newblock \emph{ACM Trans. Asian Low-Resour. Lang. Inf. Process.}, 22(4).

\bibitem[{Ushio et~al.(2023)Ushio, Zhou, and Camacho-Collados}]{ushio-etal-2023-efficient}
Asahi Ushio, Yi~Zhou, and Jose Camacho-Collados. 2023.
\newblock \href {https://doi.org/10.18653/v1/2023.findings-emnlp.981} {Efficient {M}ultilingual {L}anguage {M}odel {C}ompression through {V}ocabulary {T}rimming}.
\newblock In \emph{Findings of the Association for Computational Linguistics: EMNLP 2023}, pages 14725--14739, Singapore. Association for Computational Linguistics.

\bibitem[{Vilar and Federico(2021)}]{vilar-federico-2021-statistical}
David Vilar and Marcello Federico. 2021.
\newblock \href {https://doi.org/10.18653/v1/2021.iwslt-1.31} {A statistical extension of byte-pair encoding}.
\newblock In \emph{Proceedings of the 18th International Conference on Spoken Language Translation (IWSLT 2021)}, pages 263--275, Bangkok, Thailand (online). Association for Computational Linguistics.

\bibitem[{Wang et~al.(2024)Wang, Li, Jiang, Ding, Jiang, Liang, and Yang}]{wang2024tokenization}
Dixuan Wang, Yanda Li, Junyuan Jiang, Zepeng Ding, Guochao Jiang, Jiaqing Liang, and Deqing Yang. 2024.
\newblock \href {https://arxiv.org/pdf/2405.17067} {Tokenization {M}atters! {D}egrading {L}arge {L}anguage {M}odels through {C}hallenging {T}heir {T}okenization}.
\newblock \emph{arXiv preprint arXiv:2405.17067}.

\bibitem[{Wu et~al.(2016)Wu, Schuster, Chen, Le, Norouzi, Macherey, Krikun, Cao, Gao, Macherey, Klingner, Shah, Johnson, Liu, Kaiser, Gouws, Kato, Kudo, Kazawa, Stevens, Kurian, Patil, Wang, Young, Smith, Riesa, Rudnick, Vinyals, Corrado, Hughes, and Dean}]{wu2016googles}
Yonghui Wu, Mike Schuster, Zhifeng Chen, Quoc~V. Le, Mohammad Norouzi, Wolfgang Macherey, Maxim Krikun, Yuan Cao, Qin Gao, Klaus Macherey, Jeff Klingner, Apurva Shah, Melvin Johnson, Xiaobing Liu, Łukasz Kaiser, Stephan Gouws, Yoshikiyo Kato, Taku Kudo, Hideto Kazawa, Keith Stevens, George Kurian, Nishant Patil, Wei Wang, Cliff Young, Jason Smith, Jason Riesa, Alex Rudnick, Oriol Vinyals, Greg Corrado, Macduff Hughes, and Jeffrey Dean. 2016.
\newblock \href {http://arxiv.org/abs/1609.08144} {Google's {N}eural {M}achine {T}ranslation {S}ystem: {B}ridging the {G}ap between {H}uman and {M}achine {T}ranslation}.
\newblock Cite arxiv:1609.08144.

\bibitem[{Yamaguchi et~al.(2024)Yamaguchi, Villavicencio, and Aletras}]{yamaguchi2024empirical}
Atsuki Yamaguchi, Aline Villavicencio, and Nikolaos Aletras. 2024.
\newblock \href {https://arxiv.org/pdf/2402.10712} {An {E}mpirical {S}tudy on {C}ross-lingual {V}ocabulary {A}daptation for {E}fficient {G}enerative {LLM} {I}nference}.
\newblock \emph{arXiv preprint arXiv:2402.10712}.

\bibitem[{Yang et~al.(2022)Yang, Cui, and Chen}]{yang-etal-2022-textpruner}
Ziqing Yang, Yiming Cui, and Zhigang Chen. 2022.
\newblock \href {https://doi.org/10.18653/v1/2022.acl-demo.4} {{T}ext{P}runer: {A} {M}odel {P}runing {T}oolkit for {P}re-{T}rained {L}anguage {M}odels}.
\newblock In \emph{Proceedings of the 60th Annual Meeting of the Association for Computational Linguistics: System Demonstrations}, pages 35--43, Dublin, Ireland. Association for Computational Linguistics.

\bibitem[{Yehezkel and Pinter(2023)}]{yehezkel-pinter-2023-incorporating}
Shaked Yehezkel and Yuval Pinter. 2023.
\newblock \href {https://doi.org/10.18653/v1/2023.eacl-main.45} {Incorporating {C}ontext into {S}ubword {V}ocabularies}.
\newblock In \emph{Proceedings of the 17th Conference of the European Chapter of the Association for Computational Linguistics}, pages 623--635, Dubrovnik, Croatia. Association for Computational Linguistics.

\bibitem[{Zipf(1945)}]{zipf1945meaning}
George~Kingsley Zipf. 1945.
\newblock The meaning-frequency relationship of words.
\newblock \emph{The Journal of general psychology}, 33(2):251--256.

\bibitem[{Zouhar et~al.(2023)Zouhar, Meister, Gastaldi, Du, Sachan, and Cotterell}]{zouhar-etal-2023-tokenization}
Vil{\'e}m Zouhar, Clara Meister, Juan Gastaldi, Li~Du, Mrinmaya Sachan, and Ryan Cotterell. 2023.
\newblock \href {https://doi.org/10.18653/v1/2023.acl-long.284} {Tokenization and the noiseless channel}.
\newblock In \emph{Proceedings of the 61st Annual Meeting of the Association for Computational Linguistics (Volume 1: Long Papers)}, pages 5184--5207, Toronto, Canada. Association for Computational Linguistics.

\end{thebibliography}

\appendix

\section{Inference options}
\label{app:inference}

Picky BPE inference strictly follows the training order of events and executes merges and removals in the same chronological order (Algorithm~\ref{alg:tokenization}). Concurrent works use a different approach to inference: the input text is first tokenized with a vanilla BPE tokenizer using both active and removed tokens and then the low-frequency~\citep{cognetta2024analysis} or scaffold~\citep{lian2024scaffold} tokens are split into the shortest available sequences of valid tokens. The latter approach is suboptimal, as the training events order is likely to be broken.

For example, imagine the token sequence \texttt{[t, h, e, r, e]} on a certain training step. Tokens \texttt{(h, e)} are merged into \texttt{he} (event $e_{i_1}$). The sequence becomes \texttt{[t, he, r, e]}. Later, token \texttt{he} becomes useless and is removed (event $e_{i_2}, i_2 > i_1$). Thus, the sequence returns to \texttt{[t, h, e, r, e]}. It can happen now that tokens \texttt{(e, r)} are merged into a new token \texttt{er} (event $e_{i_3}, i_3 > i_2$). The resulting tokenization is \texttt{[t, h, er, e]}. Picky BPE tokenization will follow event order $e_{i_1},..., e_{i_2},..., e_{i_3}$ and result in \texttt{[t, h, er, e]}. The tokenization when the tokens are removed after the vanilla BPE process will first achieve \texttt{[t, he, r, e]}, as it will execute all the available merges. In a simplified example, there are no merges to perform after this step, and the algorithm will move to the removals phase: \texttt{he} will be split, and the resulting tokenization will become \texttt{[t, h, e, r, e]}. Therefore, \texttt{er} will not be merged, as it happened after the removal and contains a part of the removed token.

When repeated several times, the described issue may lead to undesired tokenization results and compromise compression. In Table~\ref{tab:inference}, we compare the compression rates of the two methods. The compression issues become more pronounced with lower thresholds as more tokens are removed.

Apart from the described inference methods, Picky BPE can use any inference method requiring a fixed vocabulary: for example, greedy left-to-right decoding~\citep{wu2016googles} or recently introduced PathPiece~\citep{schmidt2024tokenization}.

\begin{table}[t!]
    \centering
    \begin{tabular}{ccc}
    \toprule
        \multirow{2}{*}{$\mathcal{T}$} & \multirow{2}{*}{
        \begin{tabular}{c}
            BPE inference\\
            with post-removal
        \end{tabular}
        } & \multirow{2}{*}{
        \begin{tabular}{c}
            Picky BPE\\
            inference
        \end{tabular}
        } \\
        && \\
        \midrule
        1.0 & 1.000 & 1.000 \\
        \midrule
        0.9 & 0.998 & \textbf{0.997} \\
        0.8 & 0.998 & \textbf{0.996} \\
        0.7 & 1.000 & \textbf{0.994} \\
        0.6 & 1.005 & \textbf{0.992} \\
    \bottomrule
    \end{tabular}
    \caption{Comparison of compression rates $(\downarrow)$ for the vanilla BPE inference followed by splitting undesired tokens and Picky BPE inference by events order for EN--DE vocabularies of size 32768. The compression rates are shown for English.}
    \label{tab:inference}
\end{table}

\section{Training details}
\label{app:arch}

Table~\ref{tab:hyperparameters} shows the main model and training hyperparameters we used in every machine translation experiment. We trained every model for 20 epochs, except for a larger vocabulary of 32768 tokens where we trained for 25 epochs, on a single NVIDIA A40 GPU (driver version 555.42.02, CUDA version 12.5).

\begin{table}[!t]
    \centering
    \begin{tabular}{lc}
    \toprule
      Parameter & Value \\
    \midrule
      Encoder layers & 6 \\
      Decoder layers & 6 \\
      Embedding dim & 512 \\
      Hidden dim & 1024 \\
      Attention heads & 4 \\
    \midrule
      Max tokens in a batch & 4096 \\
      Optimizer & Adam \\
      Weight decay & 1e-4 \\
      Learning rate (LR) & 5e-4 \\
      LR Scheduler & inverse sqrt \\
      Warmup steps & 4000 \\
      Precision & \texttt{fp16} \\
    \bottomrule
    \end{tabular}
    \caption{\texttt{transformer-iwslt} architecture and training details configuration from \texttt{fairseq}~\cite{ott-etal-2019-fairseq}.}
    \label{tab:hyperparameters}
\end{table}

\begin{figure*}[ht!]
    \begin{subfigure}[b]{0.48\textwidth}
        \includegraphics[width=0.93\textwidth]{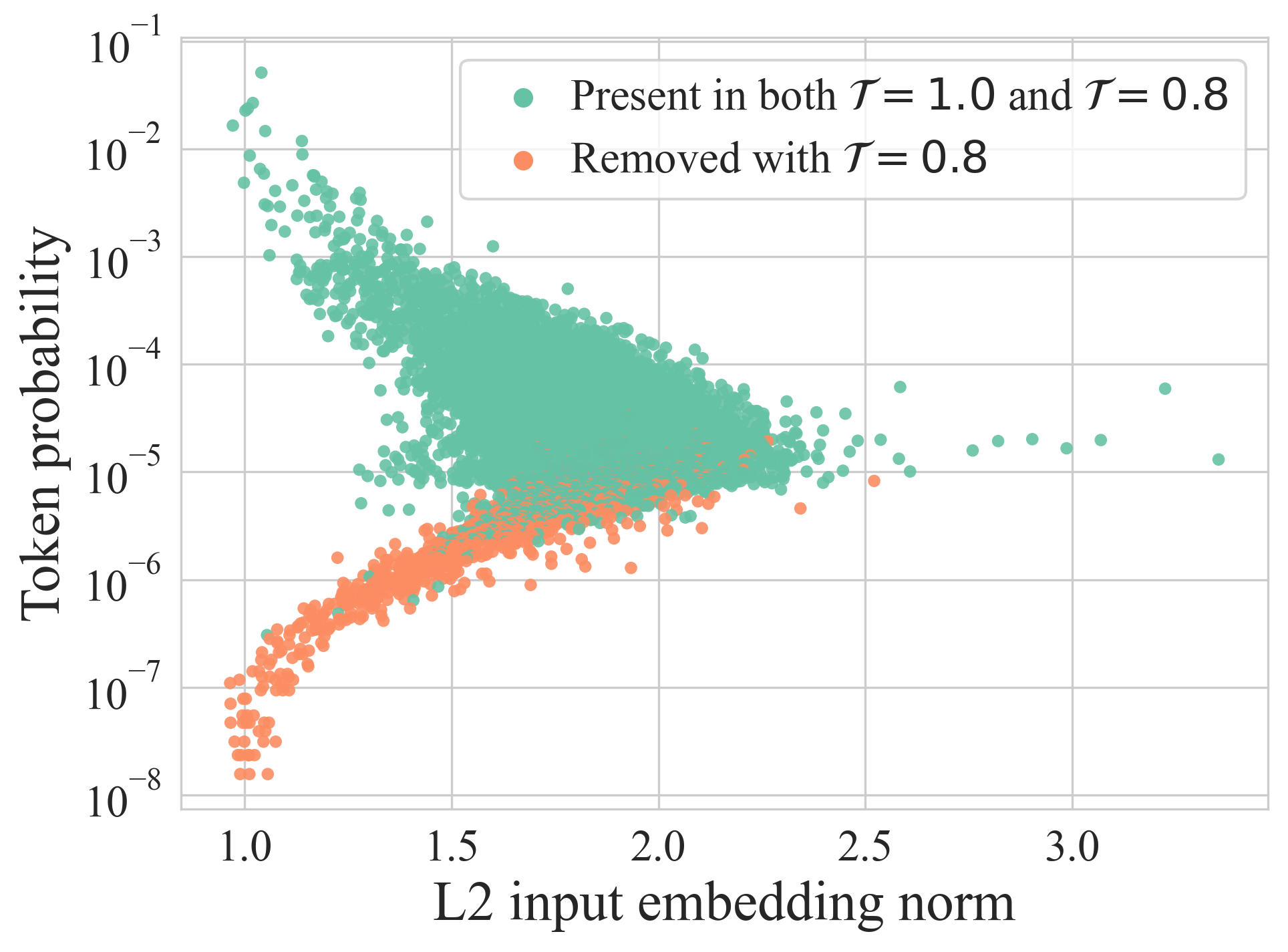}
        \subcaption{Picky BPE tokens when $\mathcal{T} = 1.0$. The tokens that are present when $\mathcal{T} = 1.0$ but are removed when $\mathcal{T} = 0.8$ (orange) are generally infrequent and have low L2 embedding norms, thus the majority of them are likely to be undertrained~\citep{land2024fishing}.}
        \label{fig:removed-0-8}
    \end{subfigure}
    \hfill
    \begin{subfigure}[b]{0.48\textwidth}
        \includegraphics[width=0.93\textwidth]{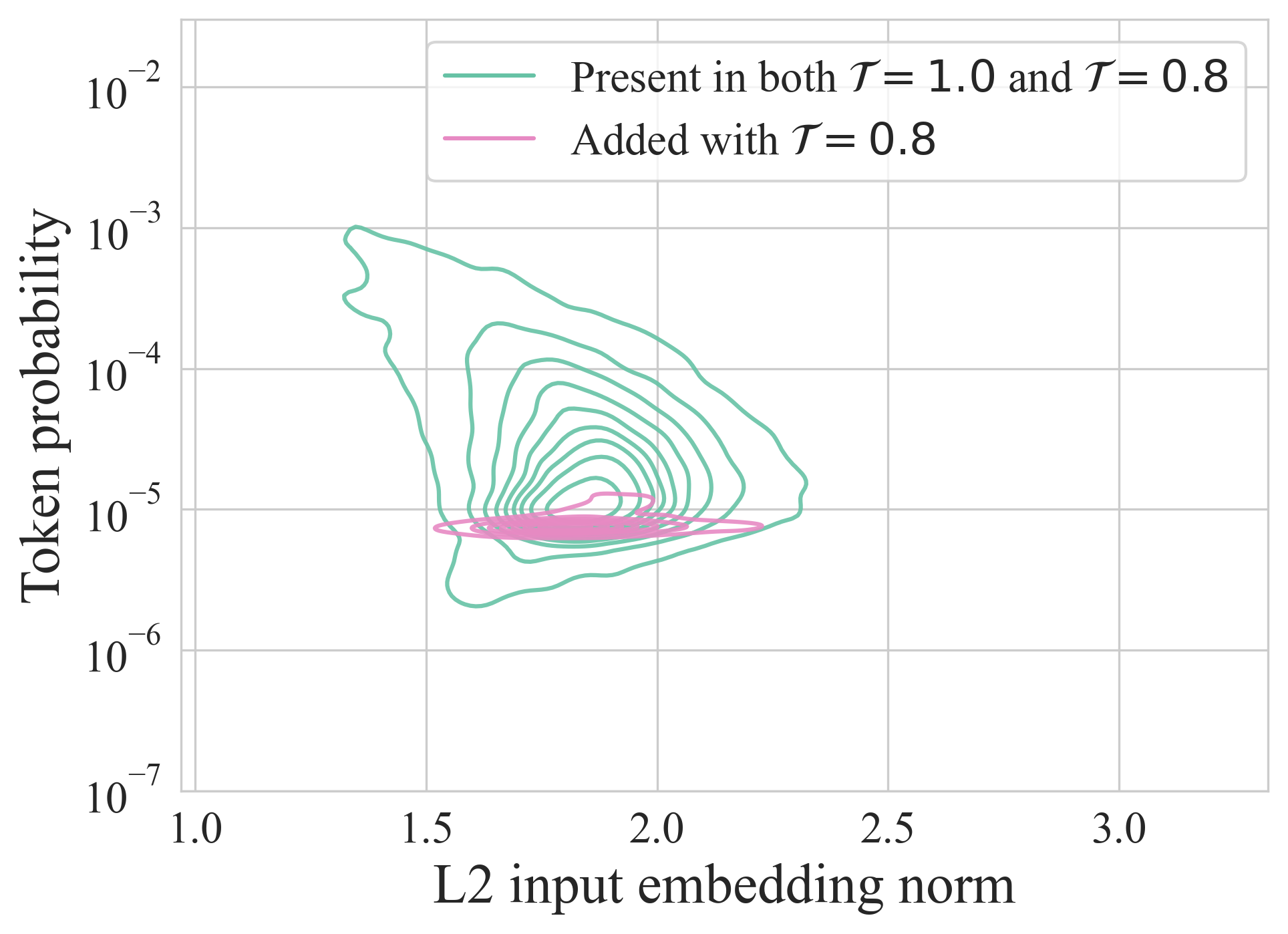}
        \subcaption{Picky BPE tokens when $\mathcal{T} = 0.8$. The tokens that are present when $\mathcal{T} = 0.8$ but not when $\mathcal{T} = 1.0$ (pink) have frequencies and L2-norms of the embeddings close to the blob center and thus are less likely to be under-trained~\citep{land2024fishing}.}
        \label{fig:added-0-8}
    \end{subfigure}
    \par\bigskip
    \begin{subfigure}[b]{0.48\textwidth}
        \includegraphics[width=0.93\textwidth]{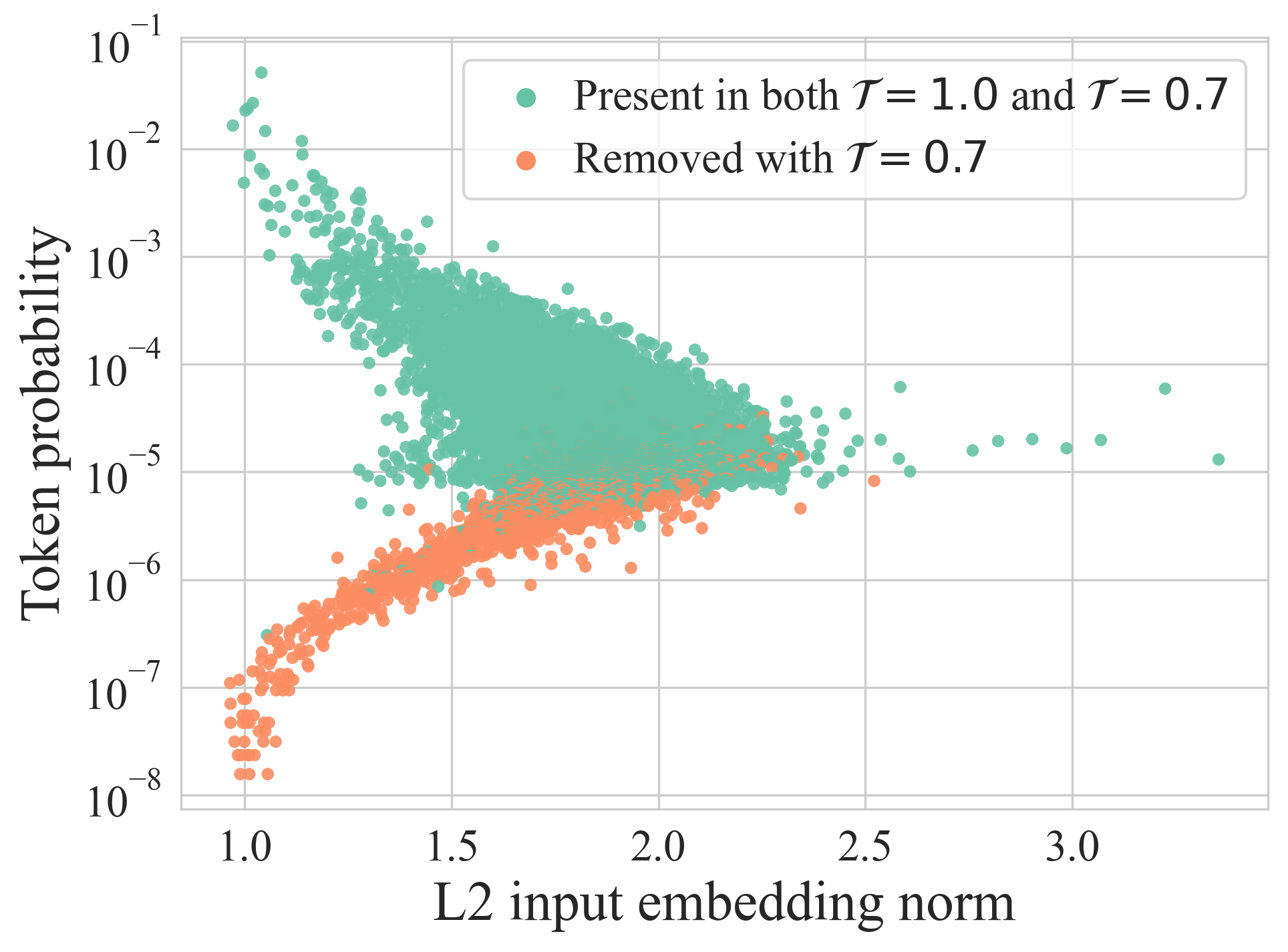}
        \subcaption{Picky BPE tokens when $\mathcal{T} = 1.0$. The tokens that are present when $\mathcal{T} = 1.0$ but are removed when $\mathcal{T} = 0.7$ (orange) are generally infrequent and have low L2 embedding norms, thus the majority of them are likely to be undertrained~\citep{land2024fishing}.}
        \label{fig:removed-0-7}
    \end{subfigure}
    \hfill
    \begin{subfigure}[b]{0.48\textwidth}
        \includegraphics[width=0.93\textwidth]{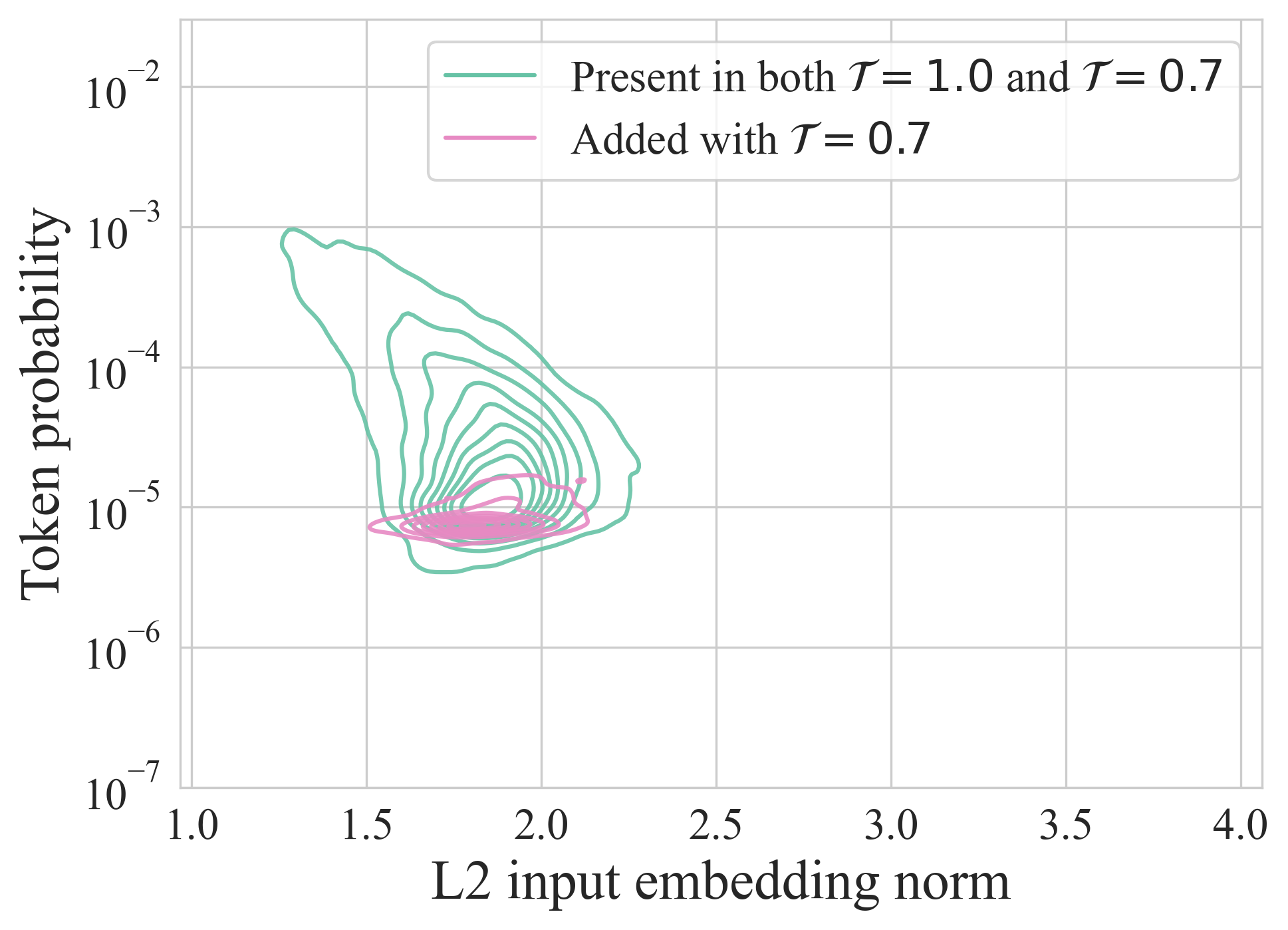}
        \subcaption{Picky BPE tokens when $\mathcal{T} = 0.7$. The tokens that are present when $\mathcal{T} = 0.7$ but not when $\mathcal{T} = 1.0$ (pink) have frequencies and L2-norms of the embeddings close to the blob center and thus are less likely to be under-trained~\citep{land2024fishing}.}
        \label{fig:added-0-7}
    \end{subfigure}
    \par\bigskip
    \begin{subfigure}[b]{0.48\textwidth}
        \includegraphics[width=0.93\textwidth]{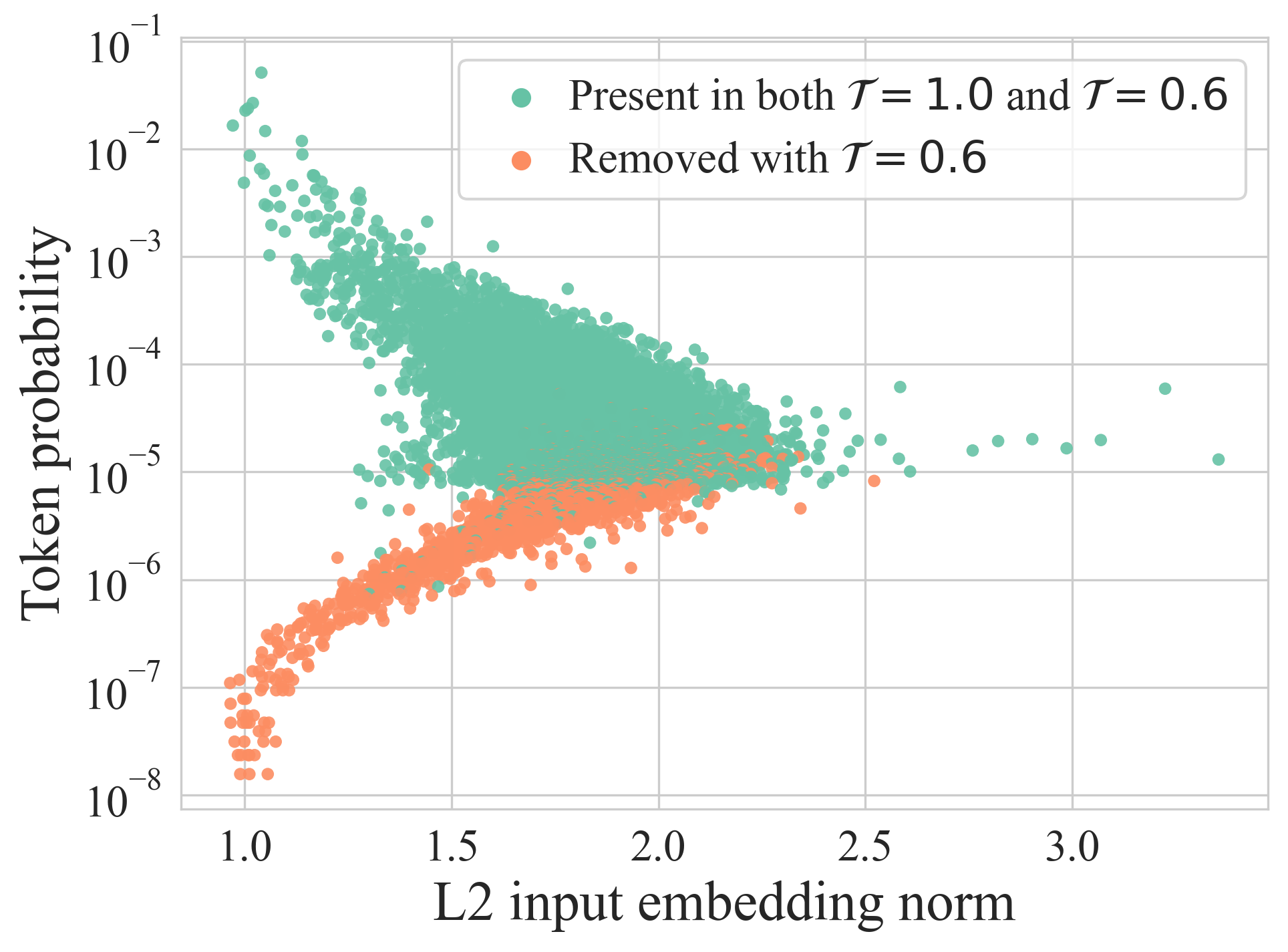}
        \subcaption{Picky BPE tokens when $\mathcal{T} = 1.0$. The tokens that are present when $\mathcal{T} = 1.0$ but are removed when $\mathcal{T} = 0.6$ (orange) are generally infrequent and have low L2 embedding norms, thus the majority of them are likely to be undertrained~\citep{land2024fishing}.}
        \label{fig:removed-0-6}
    \end{subfigure}
    \hfill
    \begin{subfigure}[b]{0.48\textwidth}
        \includegraphics[width=0.93\textwidth]{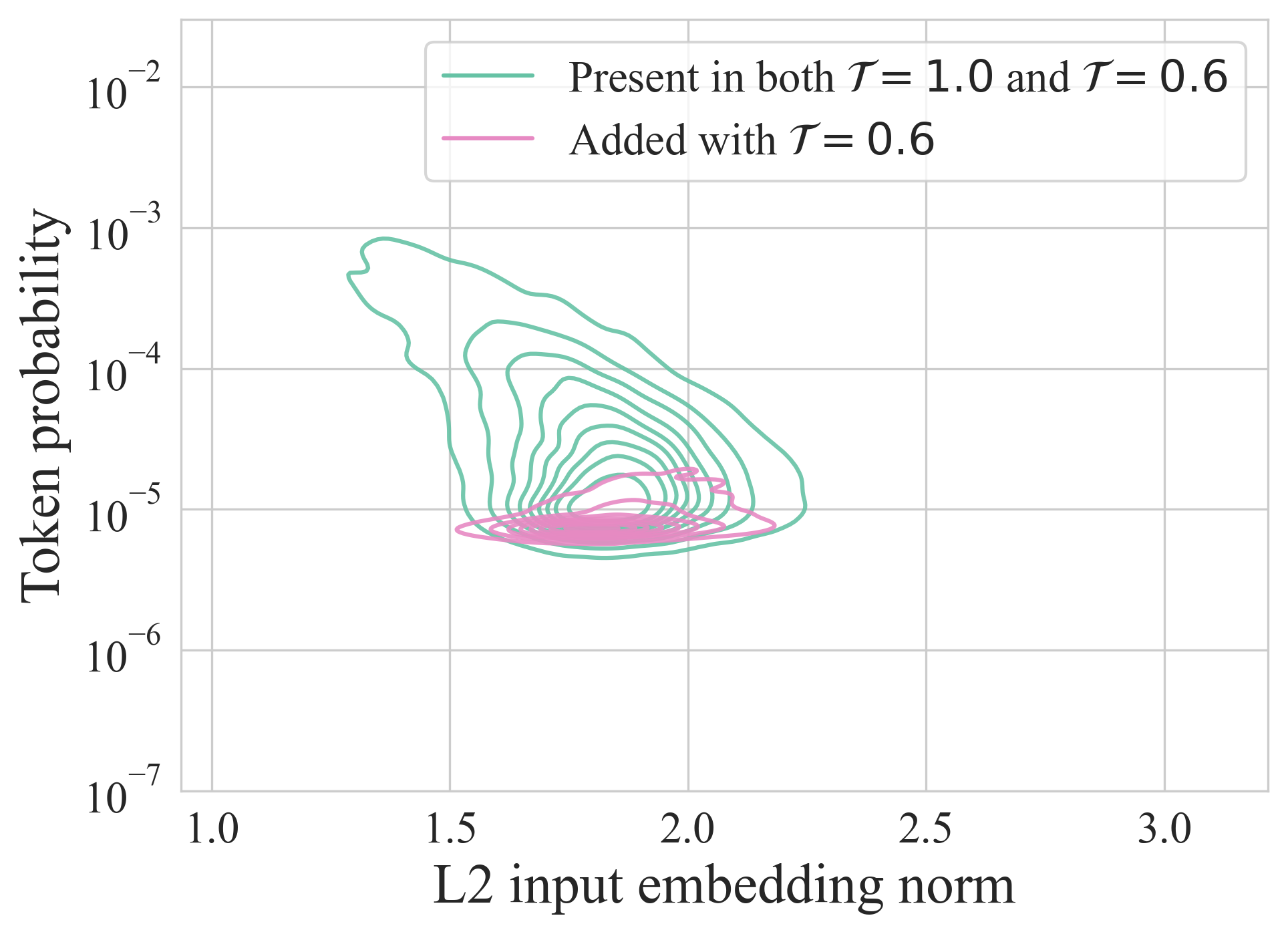}
        \subcaption{Picky BPE tokens when $\mathcal{T} = 0.6$. The tokens that are present when $\mathcal{T} = 0.6$ but not when $\mathcal{T} = 1.0$ (pink) have frequencies and L2-norms of the embeddings close to the blob center and thus are less likely to be under-trained~\citep{land2024fishing}.}
        \label{fig:added-0-6}
    \end{subfigure}
    \caption{Input embedding vectors for Picky BPE tokens with \textbf{(a, c, e)} $\mathcal{T} = 1.0$, \textbf{(b)} $\mathcal{T} = 0.8$, \textbf{(d)} $\mathcal{T} = 0.7$, and \textbf{(f)} $\mathcal{T} = 0.6$ for English vocabularies of size 16384 in EN--DE experiments with separate vocabularies. For each token we compute its probability in the training corpus (y-axis), and the L2 norm of its embedding vector in the trained model (x-axis).}
    \label{fig:oov-appendix}
\end{figure*}

\section{Under-trained tokens inspection}
\label{app:under-trained}
Figure~\ref{fig:oov-appendix} shows examples of token embedding norm distributions for thresholds 0.6, 0.7, and 0.8. As we lower the threshold, the proportion of unique tokens gets larger. However, there is no change in their nature: we remove mostly infrequent tokens and add more frequent tokens with higher norms that are close to the overall distribution.

\section{Number of Added/Removed Tokens} \label{app:removed}

Tables \ref{tab:num_rem_deen}, \ref{tab:num_rem_deet}, and \ref{tab:num_rem_uket}, report the number of added/removed tokens for each tokenizer. This is equivalent to the size of $V_i$, discussed in \S\ref{sec:algorithm_description}.

\begin{table}[h!]
\centering
\begin{tabularx}{0.95\linewidth}{ccc}\toprule
{Vocabulary} & {\multirow{2}{*}{Threshold}} & {Added / Removed} \\ 
{Size} & {} & {Tokens} \\ 
\midrule
 & 0.9 & 160 \\
 & 0.8 & 358 \\
 & 0.7 & 588 \\
\multirow{-4}{*}{8192} & 0.6 & 805 \\\midrule
 & 0.9 & 342 \\
 & 0.8 & 707 \\
 & 0.7 & 1092 \\
\multirow{-4}{*}{16384} & 0.6 & 1468 \\\midrule
 & 0.9 & 677 \\
 & 0.8 & 1280 \\
 & 0.7 & 1970 \\ 
\multirow{-4}{*}{32768} & 0.6 & 2563 \\
\midrule
 & 0.9 & 1149 \\
 & 0.8 & 2165 \\
 & 0.7 & 3312 \\
\multirow{-4}{*}{65536} & 0.6 & 4431\\ \bottomrule
\end{tabularx}
\caption{Numbers of added (removed) tokens at different thresholds for the EN--DE tokenizers used for the translation experiments.}
\label{tab:num_rem_deen}
\end{table}

\begin{table}[h!]
\centering
\begin{tabularx}{0.95\linewidth}{ccc}\toprule
{Vocabulary} & {\multirow{2}{*}{Threshold}} & {Added / Removed} \\ 
{Size} & {} & {Tokens} \\ 
\midrule
 & 0.9 & 133 \\
 & 0.8 & 313 \\
 & 0.7 & 506 \\ 
\multirow{-4}{*}{8192} & 0.6 & 718 \\
\bottomrule
\end{tabularx}
\caption{Numbers of added (removed) tokens at different thresholds for the DE--ET tokenizers used for the translation experiments.}
\label{tab:num_rem_deet}
\end{table}

\begin{table}[h!]
\centering
\begin{tabularx}{0.95\linewidth}{ccc}\toprule
{Vocabulary} & {\multirow{2}{*}{Threshold}} & {Added / Removed} \\ 
{Size} & {} & {Tokens} \\ 
\midrule
 & 0.9 & 107 \\
 & 0.8 & 255 \\
 & 0.7 & 446 \\ 
\multirow{-4}{*}{8192} & 0.6 & 605 \\
\bottomrule
\end{tabularx}
\caption{Numbers of added (removed) tokens at different thresholds for the UK--ET tokenizers used for the translation experiments.}
\label{tab:num_rem_uket}
\end{table}

\section{Compression}
\label{app:compression}

In Tables~\ref{tab:compression-ende}, \ref{tab:compression-deet}, and \ref{tab:compression-uket}, we show compression metrics for Picky BPE tokenizers relative to the vanilla BPE. We notice that compression is most pronounced in smaller vocabularies, as for the sizes of the datasets that we used larger vocabularies have large redundancy and a larger partition of tokens is allowed to be unused.

\begin{table}[h!]
\centering
\begin{tabularx}{0.9\linewidth}{cccc}
\toprule
\multirow{2.4}{*}{\shortstack{Vocabulary\\size}}&\multirow{2.4}{*}{T} & \multicolumn{2}{c}{Compression $(\downarrow)$} \\ 
 \cmidrule{3-4}
&& English & German \\
\midrule
& 1.0 & 1.000 & 1.000 \\
 & 0.9 & 0.997 & 0.996 \\
 & 0.8 & 0.995 & 0.993 \\
 & 0.7 & 0.994 & 0.991 \\
\multirow{-5}{*}{8192} & 0.6 & 0.992 & 0.989 \\
\midrule
& 1.0 & 1.000 & 1.000 \\
 & 0.9 & 0.996 & 0.998 \\
 & 0.8 & 0.994 & 0.996 \\
 & 0.7 & 0.993 & 0.995 \\ 
\multirow{-5}{*}{16384} & 0.6 & 0.991 & 0.993 \\
\midrule
& 1.0 & 1.000 & 1.000 \\
 & 0.9 & 0.997 & 0.998 \\
 & 0.8 & 0.996 & 0.998 \\
 & 0.7 & 0.994 & 0.997 \\ 
\multirow{-5}{*}{32768} & 0.6 & 0.992 & 0.996 \\
\midrule
& 1.0 & 1.000 & 1.000 \\
 & 0.9 & 0.998 & 0.998 \\
 & 0.8 & 0.997 & 0.998 \\
 & 0.7 & 0.997 & 0.998 \\ 
\multirow{-5}{*}{65536} & 0.6 & 0.996 & 0.997 \\
\bottomrule
\end{tabularx}
\caption{Compression for EN--DE tokenizers with different vocabulary sizes. The score is computed as corpus token count relative to the vanilla BPE (T = 1)}
\label{tab:compression-ende}
\end{table}

\begin{table}[h!]
\centering
\begin{tabularx}{0.9\linewidth}{cccc}
\toprule
\multirow{2.4}{*}{\shortstack{Vocabulary\\size}}&\multirow{2.4}{*}{T} & \multicolumn{2}{c}{Compression $(\downarrow)$} \\ 
 \cmidrule{3-4}
&& German & Estonian \\
\midrule
& 1.0 & 1.000 & 1.000 \\
 & 0.9 & 0.998 & 0.998 \\
 & 0.8 & 0.994 & 0.996 \\
 & 0.7 & 0.991 & 0.993 \\ 
\multirow{-5}{*}{8192} & 0.6 & 0.989 & 0.991 \\
\bottomrule
\end{tabularx}
\caption{Compression for DE--ET tokenizers with a vocabulary size of 8192. The score is computed as corpus token count relative to the vanilla BPE (T = 1)}
\label{tab:compression-deet}
\end{table}

\begin{table}[h!]
\centering
\begin{tabularx}{0.9\linewidth}{cccc}
\toprule
\multirow{2.4}{*}{\shortstack{Vocabulary\\size}}&\multirow{2.4}{*}{T} & \multicolumn{2}{c}{Compression $(\downarrow)$} \\ 
 \cmidrule{3-4}
&& Ukrainian & Estonian \\
\midrule
& 1.0 & 1.000 & 1.000 \\
 & 0.9 & 0.998 & 0.998 \\
 & 0.8 & 0.996 & 0.996 \\
 & 0.7 & 0.993 & 0.994 \\ 
\multirow{-5}{*}{8192} & 0.6 & 0.992 & 0.993 \\
\bottomrule
\end{tabularx}
\caption{Compression for UK--ET tokenizers with a vocabulary size of 8192. The score is computed as corpus token count relative to the vanilla BPE (T = 1)}
\label{tab:compression-uket}
\end{table}

\section{Word-Initial Tokens}
\label{app:word-initial}

In Tables~\ref{tab:word-initial-ende}, \ref{tab:word-initial-deet}, and \ref{tab:word-initial-uket}, we show the proportions of added and removed word-initial tokens for different vocabulary sizes and language pairs. In Tables~\ref{tab:word-initial-ende2}, \ref{tab:word-initial-deet2}, and \ref{tab:word-initial-uket2}, we show overall proportions of word-initial tokens.

\begin{table}[h!]
\centering
\begin{tabularx}{0.95\linewidth}{cccc}
\toprule
\multirow{2.4}{*}{\shortstack{Vocabulary\\size}}&\multirow{2.4}{*}{T} & \multicolumn{2}{c}{\% Word-Initial Tokens} \\ 
 \cmidrule{3-4}
&& Dropped & Added \\
\midrule
  % 1 &  &  \\ 
& 0.9 & 43.8 & 65.5 \\
 & 0.8 & 41.1 & 67.5 \\
 & 0.7 & 42.0 & 66.9 \\
\multirow{-4}{*}{8192} & 0.6 & 42.1 & 64.2 \\
\midrule
  % 1 &  &  \\ 
& 0.9 & 43.9 & 69.6 \\
 & 0.8 & 43.7 & 67.1 \\
 & 0.7 & 45.3 & 68.1 \\
\multirow{-4}{*}{16384} & 0.6 & 45.3 & 65.8 \\
\midrule
  % 1 &  &  \\ 
& 0.9 & 46.7 & 73.3 \\
 & 0.8 & 44.8 & 68.3 \\
 & 0.7 & 47.5 & 68.5 \\
\multirow{-4}{*}{32768} & 0.6 & 48.7 & 67.9 \\
\midrule
  % 1 &  &  \\ 
& 0.9 & 50.6 & 74.6 \\
 & 0.8 & 49.2 & 71.0 \\
 & 0.7 & 51.5 & 69.9 \\ 
\multirow{-4}{*}{65536} & 0.6 & 52.0 & 69.0 \\
\bottomrule
\end{tabularx}
\caption{Percent of word-initial tokens out of added and removed tokens for the
EN--DE tokenizers. Added tokens are relative to the vanilla (T = 1) tokenizer of the same vocabulary size and language pair.}
\label{tab:word-initial-ende}
\end{table}

\begin{table}[h!]
\centering
\begin{tabularx}{0.95\linewidth}{cccc}
\toprule
\multirow{2.4}{*}{\shortstack{Vocabulary\\size}}&\multirow{2.4}{*}{T} & \multicolumn{2}{c}{\% Word-Initial Tokens} \\ 
 \cmidrule{3-4}
&& Dropped & Added \\
\midrule
  % 1 &  &  \\ 
& 0.9 & 33.1 & 60.9 \\
 & 0.8 & 32.3 & 63.3 \\
 & 0.7 & 37.0 & 60.3 \\
\multirow{-4}{*}{8192} & 0.6 & 40.4 & 58.4 \\
\bottomrule
\end{tabularx}
\caption{Percent of word-initial tokens out of added and removed tokens for the
DE--ET tokenizers. Added tokens are relative to the vanilla (T = 1) tokenizer of the same vocabulary size and language pair.}
\label{tab:word-initial-deet}
\end{table}

\begin{table}[h!]
\centering
\begin{tabularx}{0.95\linewidth}{cccc}
\toprule
\multirow{2.4}{*}{\shortstack{Vocabulary\\size}}&\multirow{2.4}{*}{T} & \multicolumn{2}{c}{\% Word-Initial Tokens} \\ 
 \cmidrule{3-4}
&& Dropped & Added \\
\midrule
  % 1 &  &  \\ 
& 0.9 & 31.8 & 73.6 \\
 & 0.8 & 33.3 & 66.3 \\
 & 0.7 & 37.4 & 61.8 \\
\multirow{-4}{*}{8192} & 0.6 & 39.0 & 61.3 \\
\bottomrule
\end{tabularx}
\caption{Percent of word-initial tokens out of added and removed tokens for the
UK--ET tokenizers. Added tokens are relative to the vanilla BPE (T = 1) of the same vocabulary size and language pair.}
\label{tab:word-initial-uket}
\end{table}

\begin{table}[h!]
\centering
\begin{tabularx}{0.9\linewidth}{ccc}
\toprule
{Vocabulary} & {\multirow{2}{*}{Threshold}} & {\% Word-} \\ 
{Size} & {} & {Initial Tokens} \\ 
\midrule
   & 1.0 & 61.5 \\
 & 0.9 & 61.9 \\
 & 0.8 & 62.7 \\
 & 0.7 & 63.3 \\ 
\multirow{-5}{*}{8192} & 0.6 & 63.6 \\
\midrule
 & 1.0 & 68.0 \\
 & 0.9 & 68.6 \\
 & 0.8 & 69.2 \\
 & 0.7 & 69.7 \\ 
\multirow{-5}{*}{16384} & 0.6 & 70.0 \\
\midrule
 & 1.0 & 72.2 \\
 & 0.9 & 72.8 \\
 & 0.8 & 73.2 \\
 & 0.7 & 73.6 \\
\multirow{-5}{*}{32768} & 0.6 & 73.9 \\
\midrule
& 1.0 & 75.2 \\
 & 0.9 & 75.7 \\
 & 0.8 & 76.1 \\
 & 0.7 & 76.3 \\
\multirow{-5}{*}{65536} & 0.6 & 76.6 \\
\bottomrule
\end{tabularx}
\caption{Overall proportion of word-initial tokens at different thresholds for the EN--DE tokenizers used for the translation experiments.}
\label{tab:word-initial-ende2}
\end{table}

\begin{table}[h!]
\centering
\begin{tabularx}{0.9\linewidth}{ccc}
\toprule
{Vocabulary} & {\multirow{2}{*}{Threshold}} & {\% Word-} \\ 
{Size} & {} & {Initial Tokens} \\ 
\midrule
  & 1.0 & 58.1 \\
 & 0.9 & 58.6 \\
 & 0.8 & 59.4 \\
 & 0.7 & 59.8 \\
\multirow{-5}{*}{8192} & 0.6 & 60.0 \\
\bottomrule
\end{tabularx}
\caption{Proportion of word-initial tokens at different thresholds for the DE--ET tokenizers used for the translation experiments.}
\label{tab:word-initial-deet2}
\end{table}

\begin{table}[h!]
\centering
\begin{tabularx}{0.9\linewidth}{ccc}
\toprule
{Vocabulary} & {\multirow{2}{*}{Threshold}} & {\% Word-} \\ 
{Size} & {} & {Initial Tokens} \\ 
\midrule
 & 1.0 & 59.8 \\
 & 0.9 & 60.4 \\
 & 0.8 & 60.9 \\
 & 0.7 & 61.1 \\ 
\multirow{-5}{*}{8192} & 0.6 & 61.5 \\
\bottomrule
\end{tabularx}
\caption{Proportion of word-initial tokens at different thresholds for the UK--ET tokenizers used for the translation experiments.}
\label{tab:word-initial-uket2}
\end{table}

\section{Token Length} 
\label{app:token_len}

In Tables~\ref{tab:token_len}, \ref{tab:token_len_deet}, and \ref{tab:token_len_uket}, we show mean token lengths over different vocabulary sizes that we used in the translation experiments.

\begin{table}[h!]
\centering
\begin{tabularx}{0.95\linewidth}{ccc}\toprule
{Vocabulary} & {\multirow{2}{*}{Threshold}} & {Mean Token} \\ 
{Size} & {} & {Length (Chars)  ($\uparrow$)} \\ 
\midrule
  & 1.0 & 5.38 \\
 & 0.9 & 5.40 \\
 & 0.8 & 5.44 \\
 & 0.7 & 5.47 \\ 
\multirow{-5}{*}{8192} & 0.6 & 5.50 \\\midrule
 & 1.0 & 6.19 \\
 & 0.9 & 6.21 \\
 & 0.8 & 6.24 \\
 & 0.7 & 6.26 \\ 
\multirow{-5}{*}{16384} & 0.6 & 6.28 \\\midrule
 & 1.0 & 6.85 \\
 & 0.9 & 6.88 \\
 & 0.8 & 6.91 \\
 & 0.7 & 6.94 \\ 
\multirow{-5}{*}{32768} & 0.6 & 6.95 \\
\midrule
 & 1.0 & 7.44 \\
 & 0.9 & 7.46 \\
 & 0.8 & 7.49 \\
 & 0.7 & 7.51 \\ 
\multirow{-5}{*}{65536} & 0.6 & 7.53\\ \bottomrule
\end{tabularx}
\caption{Mean token length at different thresholds for the EN--DE tokenizers used for the translation experiments.}
\label{tab:token_len}
\end{table}

\begin{table}[h!]
\centering
\begin{tabularx}{0.95\linewidth}{ccc}\toprule
{Vocabulary} & {\multirow{2}{*}{Threshold}} & {Mean Token} \\ 
{Size} & {} & {Length (Chars)  ($\uparrow$)} \\ 
\midrule
 & 1.0 & 5.35 \\
 & 0.9 & 5.38 \\
 & 0.8 & 5.40 \\
 & 0.7 & 5.41 \\ 
\multirow{-5}{*}{8192} & 0.6 & 5.42\\
\bottomrule
\end{tabularx}
\caption{Mean token length at different thresholds for the DE--ET tokenizers used for the translation experiments.}
\label{tab:token_len_deet}
\end{table}

\begin{table}[t]
\centering
\begin{tabularx}{0.95\linewidth}{ccc}\toprule
{Vocabulary} & {\multirow{2}{*}{Threshold}} & {Mean Token} \\ 
{Size} & {} & {Length (Chars)  ($\uparrow$)} \\ 
\midrule
 & 1.0 & 4.84 \\
 & 0.9 & 4.85 \\
 & 0.8 & 4.86 \\
 & 0.7 & 4.88 \\ 
\multirow{-5}{*}{8192} & 0.6 & 4.90\\
\bottomrule
\end{tabularx}
\caption{Mean token length at different thresholds for the UK--ET tokenizers used for the translation experiments.}
\label{tab:token_len_uket}
\end{table}

\end{document}